\documentclass[sigconf, nonacm]{acmart}
\newcommand\vldbdoi{XX.XX/XXX.XX}
\newcommand\vldbpages{XXX-XXX}
\newcommand\vldbvolume{14}
\newcommand\vldbissue{1}
\newcommand\vldbyear{2023}

\newcommand\vldbavailabilityurl{URL_TO_YOUR_ARTIFACTS}
\newcommand\vldbpagestyle{plain} 
\usepackage{hyperref}
\usepackage{makecell}
\usepackage{xcolor}
\usepackage[utf8]{inputenc} % allow utf-8 input
\usepackage[T1]{fontenc}    % use 8-bit T1 fonts
\usepackage{hyperref}       % hyperlinks
\usepackage{url}            % simple URL typesetting
\usepackage{booktabs}       % professional-quality tables
\usepackage{nicefrac}       % compact symbols for 1/2, etc.
\usepackage{microtype}      % microtypography
\usepackage{xcolor}         % colors
\usepackage{soul, color, xcolor}
\soulregister{\cite}7 % 注册\cite命令
\soulregister{\citep}7 % 注册\citep命令
\soulregister{\citet}7 % 注册\citet命令
\soulregister{\ref}7 % 注册\ref命令
\soulregister{\pageref}7 % 注册\pageref命令

\usepackage{amssymb}
\usepackage{mathtools}
\usepackage{amsthm}
\usepackage{caption}
\usepackage{graphicx}
\usepackage{multirow}
\usepackage{enumitem}
\usepackage{color}
\usepackage{xcolor}
\usepackage{colortbl}
\usepackage{pifont}
\usepackage[utf8]{inputenc}
\usepackage{amsmath}  % 用于数学公式
\usepackage{xcolor}  % 用于自定义颜色
\usepackage{listings}  % 用于代码展示
\usepackage[linesnumbered,ruled,vlined]{algorithm2e}  % 用于算法显示
\usepackage{algorithmic}
\usepackage{ulem}
\usepackage{booktabs}
\usepackage{bbding}
\usepackage{wrapfig}
\usepackage{wasysym}
\usepackage{utfsym}
 
\usepackage{marvosym}
\definecolor{DarkGreen}{RGB}{1,100,32} 
\usepackage{enumitem}

\usepackage[capitalize,noabbrev]{cleveref}
\renewcommand{\thefootnote}{\fnsymbol{footnote}}
\theoremstyle{plain}
\newtheorem{theorem}{Theorem}[section]

\newtheorem{lemma}[theorem]{Lemma}

\theoremstyle{definition}
\newtheorem{definition}[theorem]{Definition}

\theoremstyle{remark}

\begin{document}
\title{OpenFGL: A Comprehensive Benchmark for Federated Graph Learning}
% \title{OpenFGL: A Comprehensive Benchmarks for Federated Graph Learning}

% \author{Xunkai Li}
% \affiliation{
% % \institution{School of Information Technology and Electrical Engineering, The University of Queensland}
% \institution{Beijing Institute of Technology}
% }
% \email{cs.xunkai.li@gmail.com}

% \author{Meihao Liao}
% \affiliation{
% % \institution{School of Information Technology and Electrical Engineering, The University of Queensland}
% \institution{Beijing Institute of Technology}
% }
% \email{mhliao@bit.edu.cn}

% \author{Zhengyu Wu}
% \affiliation{
% % \institution{School of Information Technology and Electrical Engineering, The University of Queensland}
% \institution{Beijing Institute of Technology}
% }
% \email{Jeremywzy96@outlook.com}

% \author{Wentao Zhang}
% \affiliation{
% %   \institution{Mila - Québec AI Institute, HEC Montréal}
%   \institution{Mila - Québec AI Institute \\HEC Montréal}  
% }
% \email{wentao.zhang@mila.quebec}

% \author{Rong-Hua Li}
% \affiliation{
% % \institution{School of Information Technology and Electrical Engineering, The University of Queensland}
% \institution{Beijing Institute of Technology}
% }
% \email{lironghuabit@126.com}

% \author{Guoren Wang}
% \affiliation{
% % \institution{School of Information Technology and Electrical Engineering, The University of Queensland}
% \institution{Beijing Institute of Technology}
% }
% \email{wanggrbit@126.com}

\settopmatter{authorsperrow=4} 

\author{Xunkai Li}
\affiliation{
\institution{Beijing Institute of Technology}
}
\email{cs.xunkai.li@gmail.com}

\author{Yinlin Zhu}
\affiliation{
\institution{Sun Yat-sen University}
}
\email{zhuylin27@mail2.sysu.edu.cn}

\author{Boyang Pang}
\affiliation{
\institution{Beijing Institute of Technology}
}
\email{1275854839@qq.com}

\author{Guochen Yan}
\affiliation{
  \institution{Peking University}  
}
\email{guochen_yan@outlook.com}

\author{Yeyu Yan}
\affiliation{
\institution{Beijing Jiaotong University}
}
\email{yanyeyu-work@foxmail.com}

\author{Zening Li}
\affiliation{
\institution{Beijing Institute of Technology}
}
\email{zening-li@outlook.com}

\author{Zhengyu Wu}
\affiliation{
\institution{Beijing Institute of Technology}
}
\email{Jeremywzy96@outlook.com}

\author{Wentao Zhang}
\affiliation{
  \institution{Peking University}  
}
\email{wentao.zhang@pku.edu.cn}

\author{Rong-Hua Li}
\affiliation{
\institution{Beijing Institute of Technology}
}
\email{lironghuabit@126.com}

\author{Guoren Wang}
\affiliation{
\institution{Beijing Institute of Technology}
}
\email{wanggrbit@gmail.com}

% \author{
%     {Xunkai Li\texorpdfstring{$^\dagger$},, 
%     Yinlin Zhu\texorpdfstring{$^\ddagger$},,
%     Boyang Pang\texorpdfstring{$^\dagger$},,
%     Guochen Yan\texorpdfstring{$^\diamondsuit$},,\texorpdfstring{\\}
%     ZZening Li\texorpdfstring{$^{\dagger}$},, 
%     Yeyu Yan\texorpdfstring{$^{\sharp}$},,
%     Zhengyu Wu\texorpdfstring{$^{\dagger}$},,
%     Wentao Zhang\texorpdfstring{$^{\diamondsuit}$},,
%     Rong-Hua Li\texorpdfstring{$^\dagger$},, 
%     Guoren Wang\texorpdfstring{$^\dagger$},\texorpdfstring{\\},}
%     {\texorpdfstring{$^\dagger$} BBeijing Institute of Technology}
%     {\texorpdfstring{$^\ddagger$} SSun Yat-sen University}\\
%     {\texorpdfstring{$^\diamondsuit$} BBeijing Jiaotong University}
%     {\texorpdfstring{$^\sharp$} PPeking University}\\
%     \Letter\;\;Primary contact: cs.xunkai.li@gmail.com, zhuylin27@mail2.sysu.edu.cn
% }


\begin{abstract}
    Federated graph learning (FGL) is a promising distributed training paradigm for graph neural networks across multiple local systems without direct data sharing.
    This approach inherently involves large-scale distributed graph processing, which closely aligns with the challenges and research focuses of graph-based data systems.
    Despite the proliferation of FGL, the diverse motivations from real-world applications, spanning various research backgrounds and settings, pose a significant challenge to fair evaluation.
    To fill this gap, we propose OpenFGL, a unified benchmark designed for the primary FGL scenarios: Graph-FL and Subgraph-FL.
    Specifically, OpenFGL includes 42 graph datasets from 18 application domains, 8 federated data simulation strategies that emphasize different graph properties, and 5 graph-based downstream tasks.
    Additionally, it offers 18 recently proposed SOTA FGL algorithms through a user-friendly API, enabling a thorough comparison and comprehensive evaluation of their effectiveness, robustness, and efficiency.
    Our empirical results demonstrate the capabilities of FGL while also highlighting its potential limitations, providing valuable insights for future research in this growing field, particularly in fostering greater interdisciplinary collaboration between FGL and data systems.

\end{abstract}

\maketitle

\pagestyle{\vldbpagestyle}
\begingroup\small\noindent\raggedright\textbf{PVLDB Reference Format:}\\
% \vldbauthors.
Xunkai Li, Yinlin Zhu, Boyang Pang, Guochen Yan, Zening Li, Yeyu Yan, Zhengyu Wu, Wentao Zhang, Rong-Hua Li, Guoren Wang.
OpenFGL: A Comprehensive Benchmark for Federated Graph Learning. PVLDB, \vldbvolume(\vldbissue): \vldbpages, \vldbyear.
\href{https://doi.org/\vldbdoi}{doi:\vldbdoi}
\endgroup
\begingroup
\renewcommand\thefootnote{}\footnote{\noindent
This work is licensed under the Creative Commons BY-NC-ND 4.0 International License. Visit \url{https://creativecommons.org/licenses/by-nc-nd/4.0/} to view a copy of this license. For any use beyond those covered by this license, obtain permission by emailing \href{mailto:info@vldb.org}{info@vldb.org}. Copyright is held by the owner/author(s). Publication rights licensed to the VLDB Endowment. \\
\raggedright Proceedings of the VLDB Endowment, Vol. \vldbvolume, No. \vldbissue\ %
ISSN 2150-8097. \\
\href{https://doi.org/\vldbdoi}{doi:\vldbdoi} \\
}\addtocounter{footnote}{-1}\endgroup

\ifdefempty{\vldbavailabilityurl}{}{
\begingroup\small\noindent\raggedright\textbf{PVLDB Artifact Availability:}\\
The source code, data, and/or other artifacts have been made available at {\url{https://github.com/xkLi-Allen/OpenFGL}}.
\endgroup
}


\section{Introduction}
    Recently, graphs have emerged as effective tools for capturing intricate interactions among real-world entities, leading to their widespread applications.
    Based on this, we can translate various business applications from industrial scenarios into different graph-based downstream tasks from the machine learning perspective.
    To generate effective graph entity embeddings, graph neural networks (GNNs) utilize relational data stored from databases, encoding both node features and structural information for various data systems applications~\cite{besta2022gnn_database1, tian2023gnn_database2, zhou2023gnn_database3}.
    This paradigm has been widely validated, including node-level financial fraud detection~\cite{hyun2023app_gnn_fina2, qiu2023app_gnn_fina3}, link-level recommendation~\cite{10.1007/s00521-022-07735-y_EHGCN, yang2023app_gnn_rec2}, and graph-level bioinformatics~\cite{qu2023app_gnn_bio2, gao2023app_gnn_bio3}.

    Despite their effectiveness, privacy regulations and scalability issues pose challenges to direct data sharing, complicating centralized model training~\cite {yuan2021fl_cite1,li2024fl_cite2,yuan2024fl_cite3}.
    To address this challenge, federated graph learning (FGL) has been proposed to enable collaborative training across multiple local systems~\cite{pan2023fl_cite4,gu2023fl_cite5, zhang2024fl_cite6,yao2023fl_cite7,wu2023fl_cite8}, providing a novel distributed approach to graph-based data management~\cite{sheth1990_federated_database_1,yang2019_federated_database_4,zhu2021federated_database_6,antunes2022_federated_database_8,liu2022_federated_database_9}.
    Existing FGL benchmarks, such as FS-G \cite{WangFedScope_22_fsg} (Year: 2022) and FedGraphNN \cite{he2021fedgraphnn} (Year: 2021), offer valuable insights but still have the following limitations:
    (1) \textbf{Datasets}: Limited to few application domains (e.g., citation networks and recommendation).
    (2) \textbf{Algorithms}: Missing recent SOTA FGL methods (e.g., 8 methods in 2023, 10+ methods in 2024).
    (3) \textbf{Experiments}: Lack of graph-oriented federated data simulation strategies, inadequate support for various graph-based downstream tasks, and limited evaluation perspectives.
    While the research prospects and enthusiasm for FGL are prominent and growing~\cite{wu2024fl_cite9,liang2023fl_cite10,wang2022fl_cite11,jiang2024fl_cite12}, the absence of a comprehensive benchmark for fair comparison impedes its development.
    Specifically, the diversity of downstream tasks (i.e., node, link, and graph), the unique graph properties (i.e., feature, label, and topology), and the complexity of FGL evaluation (i.e., effectiveness, robustness, and efficiency) collectively pose significant obstacles to achieving a comprehensive understanding of the current FGL landscape.
    Consequently, there is an emergency need to develop a standardized benchmark.

    In this paper, we propose OpenFGL, which integrates 2 commonly used FGL scenarios, 42 datasets in 18 application domains, 8 graph-specific distributed data simulation strategies, 18 recently proposed SOTA algorithms, and 5 graph-based downstream tasks.
    These components are implemented with a unified API to facilitate fair comparisons and future development in a user-friendly manner.
    Based on this foundation, we provide a comprehensive evaluation of existing FGL algorithms, drawing the following valuable insights.
    For \textbf{Effectiveness} we advocate for quantifying the statistics in distributed graphs to define the graph-based federated heterogeneity formally.
    For \textbf{Robustness}, to facilitate the practical deployment of existing FGL algorithms, we emphasize the significant potential of personalized, multi-client collaboration, and privacy-preserving techniques in addressing challenges such as data noise, data sparsity, low client participation, and generalization in complex applications.
    For \textbf{Efficiency}, considering the industry-scale datasets, we encourage FGL developers to prioritize algorithmic scalability and propose innovative federated collaborative paradigms that bring substantial benefits in improving efficiency.
    To further illustrate the advantages of our proposed OpenFGL compared to existing FGL benchmarks, we provide a clear description in Table~\ref{tab: benchmark comparison}.

    \textbf{Our contributions}.
    (1) \textit{\underline{Comprehensive Benchmark}}. 
    We propose OpenFGL, which integrates 2 scenarios with 42 publicly datasets.
    Based on this, we propose 8 practical distributed settings from the perspective of data heterogeneity.
    Meanwhile, we integrate 18 FGL algorithms and advocate 3 orthogonal evaluation perspectives to establish comprehensive baselines (See Table~\ref{tab: benchmark comparison}).
    We believe that FGL can inspire new research directions within the data systems community, particularly in scalable, privacy-preserving, and distributed data processing.
    (2) \textit{\underline{Valuable Insights}}.
    Leveraging the user-friendly API integrated into OpenFGL, we conducted extensive empirical studies and derived 10 valuable conclusions (See Sec.~\ref{sec: Performance Comparison}-Sec.~\ref{sec: Efficiency Evaluation}). 
    Building upon these findings, we provide 6 key insights from the perspectives of effectiveness, robustness, and efficiency, outlining promising research directions for the future FGL community (See Sec.~\ref{sec: Conclusion and Future Directions}).
    (3) \textit{\underline{Open-sourced Library and Detailed Repository}}. 
    We develop an easy-to-use and open-source library to support ongoing FGL studies, allowing users to evaluate their algorithms or datasets with ease.
    Additionally, we conduct a comprehensive review of existing FGL studies and release a beginner-friendly repository to facilitate the growth of the FGL community.
    The code and related tutorial are available at \url{https://github.com/xkLi-Allen/OpenFGL}.

\section{Problem Statement}
    In this section, we briefly review the FGL training pipeline in the following 2 most representative scenarios. 
    To begin with, from a data perspective, each client regards graphs (Graph-FL) and nodes in a subgraph (Subgraph-FL) as data samples.
    Subsequently, FGL aims to achieve collaborative training based on these clients and a trusted server. 
    Formally, we consider a FGL system consisting of $K$ clients, where the $k$-th client manages a private dataset denoted as $\mathcal{D}^{(k)} = \{\mathcal{G}_{i}^{(k)}\}_{i=1}^{N_T}$.
    Here, $N_T$ is the task-specific description, where $N_T$ denotes the number of graph samples under Graph-FL, while $N_T=1$ exists under Subgraph-FL.
    To provide a detailed description, we take FedAvg \cite{mcmahan2017fedavg} as an example. 
    Its training process within the $T$ communication round is outlined in four key steps:

    \noindent \textbf{1. Receive Message.}
    Each client initializes its local model with the unified parameters from the server at the $t$-th round $\mathbf{W}^{T}_k \leftarrow \widehat{\mathbf{W}}^{T}$;
    
    \noindent \textbf{2. Local Update.}
    Each client performs local training using its private data, i.e., $\min_{\mathbf{W}^T_k} \mathcal{L}_{task}(\mathcal{D}^{(k)})$ to obtain $\mathbf{W}^{T+1}_k$, where $\mathcal{L}_{task}$ denotes the task-specific optimization objectives.
    
    \noindent \textbf{3. Upload Message.}
    Each client uploads their local updated models $\mathbf{W}^{T+1}_k$ and the number of data samples $D_i$ (i.e., graphs, nodes, or edges, depending on the downstream task) to the server.

    \noindent \textbf{4. Global Aggregation.}
    The server aggregates the updated models to obtain $\widehat{\mathbf{W}}^{T+1}$ for the next communication, i.e., $\widehat{\mathbf{W}}^{T+1} \!\!\leftarrow \frac{1}{D} \sum_{k=1}^K D_k \mathbf{W}^{T+1}_k$, where $D$ is the total number of data samples.

% \usepackage{multirow}
\begin{table*}[]
\caption{FGL benchmark comparison, where D.D. denotes dataset domain.
Effe. and Effi. represent effectiveness and efficiency.}
\small
\label{tab: benchmark comparison}
\begin{tabular}{c|c|c|c|c|c|c}
\midrule[0.3pt]
FGL Benchmarks & D.D. & FGL Algorithms       & Federated Data Simulation                                                                       & Tasks & Evaluations     & Conclusions                                                                                          \\ \midrule[0.3pt]
FS-G~\cite{WangFedScope_22_fsg}           & 7    & 3 (Year: 2021)       & Label, Topology                                                                                 & 3     & Effe. and Effi. & 5 (Effe.+Hyperparameter)                                                                                            \\
FedGraphNN~\cite{he2021fedgraphnn}     & 6    & 0 (FL+GNNs)        & LDA-based Feature, Label                                                                        & 3     & Effe. and Effi. & 5 (Effe.+Effi.+Security)                                                                                            \\ \midrule[0.3pt]
OpenFGL (Ours) & 18   & 18 (Year: 2021-2024) & \begin{tabular}[c]{@{}c@{}}Cross-domain and Graph-based\\ Feature, Label, Topology\end{tabular} & 5     & Table~\ref{tab: An overview of OpenFGL}         & \begin{tabular}[c]{@{}c@{}}10 (Effe. + Robustness + Effi.)\\ and 6 Promising Directions\end{tabular} \\ \midrule[0.3pt]
\end{tabular}
\end{table*}

\begin{table*}[t]
\caption{An overview of OpenFGL.}
\label{tab: An overview of OpenFGL}
\begin{tabular}{c|c|c}
\midrule[0.3pt]
Data                    & \cellcolor{blue!15}{{\textit{\textbf{Graph-FL Scenario}}}}                                                   & \cellcolor{blue!15}{{\textit{\textbf{Subgraph-FL Scenario}}}}                                                                    \\ \midrule[0.3pt]
Datasets             & Protein, Collaboration, Movie Network...         & Citation, Purchase, Wiki, Syntax Network...                              \\
Simulation & Feature, Label, Topology, Cross-domain                     & Feature, Label, Community, Cross-domain                                               \\
Tasks                & Graph Regression, Graph Classification                       & Node Classification, Link Prediction, Node Clustering                                 \\ \midrule[0.3pt]
\textcolor{red}{Method}                     & \multicolumn{2}{c}{\cellcolor{red!15}{{\textit{\textbf{Algorithms}}}}}                                                                                                                     \\ \midrule[0.3pt]
GNN                  & \multicolumn{2}{c}{GCN, GAT, GraphSAGE, SGC, GCNII, GIN, TopKPooling, SAGPooling, EdgePooling, PANPooling}                                                                                           \\
FL                   & \multicolumn{2}{c}{FedAvg, FedProx, Scaffold, FedDC, MOON, FedProto, FedNH, FedTGP}                                                                       \\
FGL                  & \multicolumn{2}{c}{GCFL+, FedStar, FedSage+, Fed-PUB, FedGTA, FGSSL, FedGL, AdaFGL, FGGP, FedDEP, FedTAD}                                                           \\ \midrule[0.3pt]
\textcolor[rgb]{0.0, 0.5, 0.0}{Experiment}                      & \multicolumn{2}{c}{\cellcolor{green!15}{{\textit{\textbf{Evaluations}}}}}                                                                                                                    \\ \midrule[0.3pt]
Data Analysis    & \multicolumn{2}{c}{Feature KL Divergence, Label Distribution, Topology Statics}      \\
Effectiveness        & \multicolumn{2}{c}{MSE, RMSE, Accuracy, Precision, Recall, F1, AUC-ROC, AP, Clustering-accuracy, NMI, ARI}                                                 \\
Robustness           & \multicolumn{2}{c}{Noise, Sparsity, Client Active Fraction, Federated Scenario Generalization, DP-based Privacy Preserve} \\
Efficiency           & \multicolumn{2}{c}{Convergence, Scalability, Communication, FLOPS, Time\&Space Complexity}                                                                                                                                                                             \\ \midrule[0.3pt]
\end{tabular}
\end{table*}

\section{Benchmark Design}
\subsection{Data-level FGL Scenarios}
\label{sec: Data-level FGL Scenarios}
    In this section, we distinguish FGL scenarios based on the types of downstream tasks and the storage forms of local data at each client, categorizing them as Graph-FL, Subgraph-FL, and Node-FL. 
    This serves as the guideline for proposing the concept of data-level FGL scenarios, emphasizing the data-centric description of real-world FGL applications.
    Notably, this section focuses on the pre-experiment preparation from a data perspective, whereas Sec.~\ref{sec: Experiment-level FGL Evaluations} emphasizes a more in-depth empirical analysis through a comprehensive evaluation across 3 orthogonal perspectives.
    In OpenFGL, we focus on the two prevalent FGL scenarios:
    (1) {Graph-FL}.
    The growing integration with graph-based techniques and AI4Science applications, such as drug discovery, has motivated this scenario, in which clients consider graphs as the data samples and pursue collaborative training between clients to acquire powerful models while preserving data privacy.
    (2) {Subgraph-FL}. 
    Realistic applications in this scenario include node-level fraud detection for financial security and link-level user-item interactions for recommendation, with data stored in a distributed manner.
    Clients treat their data as subgraphs of a larger and more comprehensive global graph and focus on utilizing nodes and edges as data samples for training. 
    Due to regulatory constraints, clients seek a collaborative training scheme to develop well-trained models without direct data sharing.

    Notably, Node-FL is also a significant paradigm of FGL, which is widely used in graph-based spatial-temporal analysis, such as sensor networks~\cite{fedgnn_sensor} and traffic flow prediction~\cite{fedgnn_traffic}, where nodes are only aware of their local context within the broader network. 
    Although Node-FL has been widely mentioned, we have not integrated it into OpenFGL. 
    This is because most Node-FL studies are tailored to specific scenarios and involve experimental setups that are highly diverse and closely aligned with particular application contexts. 
    These characteristics make Node-FL less suitable for being included in a unified benchmark evaluation, where consistency across scenarios is essential for comprehensive comparisons.

    {\textbf{Datasets.}} 
    To comprehensively evaluate existing FGL algorithms, we have compiled a substantial collection of public datasets from various domains.
    Specifically, Regarding {Graph-FL} scenario, we conduct experiments on the compounds networks (MUTAG, BZR, COX2, DHFR, PTC-MR, AIDS, NCI1, hERG, ogbg-molhiv, ogbg-molpcba)~\cite{debnath1991_MUTAG, sutherland2003_COX2_BZR_DHFR, helma2001_PTC_PROTEINS, riesen2008_AIDS, wale2008_NCI1,gaulton2017_hERG,hu2020ogb}, protein networks (ENZYMES, DD, PROTEINS, ogbg-ppa)~\cite{borgwardt2005_ENZYMES, dobson2003_DD, helma2001_PTC_PROTEINS, hu2020ogb}, collaboration network (COLLAB)~\cite{leskovec2005graphs_COLLAB}, movie network (IMDB-B/M) ~\cite{yanardag2015_IMDB_B_M}, super-pixel networks (MNISTSuperPixels)~\cite{monti2017geometric}, point cloud (ShapeNet)~\cite{yi2016scalable}, and syntax trees (ogbg-code2)~\cite{hu2020ogb}.
    As for {Subgraph-FL} scenario, we perform experiments on the citation networks (Cora, Citeseer, PubMed, FedDBLP, ogbn-arxiv)~\cite{Yang16cora, WangFedScope_22_fsg, hu2020ogb}, co-purchase networks (Amazon-Computers, Amazon-Photo, ogbn-products)~\cite{shchur2018amazon_datasets, hu2020ogb}, and co-author networks (CS, Physics)~\cite{shchur2018amazon_datasets}, wiki-page networks (Chameleon, Squirrel)~\cite{pei2020geomgcn, platonov2023hete_gnn_survey4}, actor network (Actor)~\cite{pei2020geomgcn}, game synthetic network (Minesweeper)~\cite{platonov2023hete_gnn_survey4}, crowd-sourcing network (Tolokers)~\cite{platonov2023hete_gnn_survey4}, syntax network (Roman-empire)~\cite{platonov2023hete_gnn_survey4}, rating network (Amazon-rating)~\cite{platonov2023hete_gnn_survey4}, social network (Questions)~\cite{platonov2023hete_gnn_survey4}, and point cloud networks (PCPNet, S3DIS)~\cite{guerrero2018pcpnet,armeni20163d}. 

    Remarkably, besides these datasets being collected across various application domains, they exhibit diverse graph characteristics, encompassing rich or poor node attributes at the feature level, homophily or heterophily, and sparsity or density at the topology level. 
    These graph properties facilitate the evaluation of the adaptability and robustness of existing FGL algorithms across various and intricate experimental settings, highlighting their strengths and revealing potential limitations from a data-centric perspective.
    More details can be found in Tables.~\ref{tab: graph_fl_datasets},~\ref{tab: subgraph_fl_datasets} and ~\cite{OpenFGL} (\ref{appendix: Dataset Description}).

    \begin{table*}[htbp]
\setlength{\abovecaptionskip}{0.2cm}
\caption{The statistical information of Graph-FL datasets.}
\label{tab: graph_fl_datasets}
\resizebox{\linewidth}{40mm}{
\setlength{\tabcolsep}{4.8mm}{
\begin{tabular}{cccccccc}
\midrule[0.3pt]
Graph-FL    & Graphs & Nodes  & Edges    & Features & Classes & Train/Val/Test & Description           \\ \midrule[0.3pt]
MUTAG       & 188    & 17.93  & 19.79    & 7        & 2       & 80\%/10\%/10\% & Compounds Network     \\
BZR         & 405    & 35.75  & 38.36    & 56       & 2       & 80\%/10\%/10\% & Compounds Network     \\
COX2        & 467    & 41.22  & 43.45    & 38       & 2       & 80\%/10\%/10\% & Compounds Network     \\
DHFR        & 467    & 42.43  & 44.54    & 56       & 2       & 80\%/10\%/10\% & Compounds Network     \\
PTC-MR      & 344    & 14.29  & 14.69    & 18       & 2       & 80\%/10\%/10\% & Compounds Network     \\
AIDS        & 2,000  & 15.69  & 16.20     & 42       & 2       & 80\%/10\%/10\% & Compounds Network     \\
NCI1        & 4,110  & 29.87  & 32.30     & 37       & 2       & 80\%/10\%/10\% & Compounds Network     \\
hERG        &10,572  & 29.39  & 94.09     & 8       & -       & 80\%/10\%/10\% & Compounds Network     \\
ogbg-molhiv &41,127  & 25.50  & 27.50     & 9       & 2       & 80\%/10\%/10\% & Compounds Network     \\
ogbg-molpcba&437,929 & 26.00  & 28.10     & 9       & 2       & 80\%/10\%/10\% & Compounds Network     \\\midrule[0.3pt]
ENZYMES     & 600    & 32.63  & 62.14    & 21       & 6       & 80\%/10\%/10\% & Protein Network       \\
DD          & 1,178  & 284.32 & 715.66   & 89       & 2       & 80\%/10\%/10\% & Protein Network       \\
PROTEINS    & 1,113  & 39.06  & 72.82    & 4        & 2       & 80\%/10\%/10\% & Protein Network       \\
ogbg-ppa    & 158,100& 243.40  & 2,266.10 & 4        & 37       & 49\%/29\%/22\% & Protein Network       \\\midrule[0.3pt]
COLLAB      & 5,000  & 74.49  & 2,457.78 & degree   & 3       & 80\%/10\%/10\% & Collaboration Network \\
IMDB-BINARY & 1,000  & 19.77  & 96.53    & degree   & 2       & 80\%/10\%/10\% & Movie Network         \\
IMDB-MULTI  & 1,500  & 13.00     & 65.94    & degree   & 3       & 80\%/10\%/10\% & Movie Network         \\ \midrule[0.3pt]
ShapeNet & 16,881 &2616.20 &     KNN  &  3       &   50     & 40\%/40\%/40\% &  Point Cloud Network       \\
MNISTSuperPixels  & 70,000 &   75.00 & 1393.03    &  1          &       10 &  43\%/43\%/14\%&  Super-pixel Network         \\
ogbg-code2  & 452,741	 &   125.20	 & 124.20    &  4          &       275 &  90\%/5\%/5\%&  Abstract Syntax Trees         \\
\midrule[0.3pt]
\end{tabular}
}}
\vspace{-0.2cm}
\end{table*}

\begin{table*}[htbp]
\setlength{\abovecaptionskip}{0.2cm}
\caption{The statistical information of Subgraph-FL datasets.
}
\label{tab: subgraph_fl_datasets}
\resizebox{\linewidth}{44mm}{
\setlength{\tabcolsep}{5.4mm}{
\begin{tabular}{ccccccc}
\midrule[0.3pt]
Subgraph-FL       & Nodes     & Edges      & Features & Classes & Train/Val/Test & Description            \\ \midrule[0.3pt]
Cora              & 2,708     & 5,429      & 1,433    & 7       & 20\%/40\%/40\% & Citation Network       \\
CiteSeer          & 3,327     & 4,732      & 3,703    & 6       & 20\%/40\%/40\% & Citation Network       \\
PubMed            & 19,717    & 44,338     & 500      & 3       & 20\%/40\%/40\% & Citation Network       \\
FedDBLP         & 52,202   & 271,054    & 1,639      & 4      & 50\%/20\%/30\% & Citation Network       \\
ogb-arxiv         & 169,343   & 231,559    & 128      & 40      & 60\%/20\%/20\% & Citation Network       \\ \midrule[0.3pt]
Amazon-Photo      & 7,487     & 119,043    & 745      & 8       & 20\%/40\%/40\% & Co-purchase Network    \\
Amazon-Computers  & 13,381    & 245,778    & 767      & 10      & 20\%/40\%/40\% & Co-purchase Network    \\
ogb-products      & 2,449,029 & 61,859,140 & 100      & 47      & 10\%/5\%/85\%  & Co-purchase Network    \\ \midrule[0.3pt]
Co-author CS      & 18,333    & 81,894     & 6,805    & 15      & 20\%/40\%/40\% & Co-author Network      \\
Co-author Physics & 34,493    & 247,962    & 8,415    & 5       & 20\%/40\%/40\% & Co-author Network      \\
Chameleon         & 2,277     & 36,101     & 2,325    & 5       & 48\%/32\%/20\% & Wiki-page Network      \\
Chameleon Filter  & 890       & 13,584     & 2,325    & 5       & 48\%/32\%/20\% & Wiki-page Network      \\
Squirrel          & 5,201     & 216,933    & 2,089    & 5       & 48\%/32\%/20\% & Wiki-page Network      \\
Squirrel Filter   & 2,223     & 65,718     & 2,089    & 5       & 48\%/32\%/20\% & Wiki-page Network      \\ \midrule[0.3pt]
Actor             & 7,600     & 29,926     & 931      & 5       & 50\%/25\%/25\%   & Actor Network          \\
Minesweeper       & 10,000    & 39,402     & 7        & 2       & 50\%/25\%/25\%   & Game Synthetic Network \\
Tolokers          & 11,758    & 519,000    & 10       & 2       & 50\%/25\%/25\%   & Crowd-sourcing Network \\
Roman-empire      & 22,662    & 32,927     & 300      & 18      & 50\%/25\%/25\%   & Article Syntax Network        \\
Amazon-ratings     & 24,492    & 93,050     & 300      & 5       & 50\%/25\%/25\%   & Rating Network         \\
Questions         & 48,921    & 153,540    & 301      & 2       & 50\%/25\%/25\%   & Social Network         \\ \midrule[0.3pt]
PCPNet    & 100,000 &    KNN  &     5    &     -   & 26\%/10\%/64\% & Point Cloud Network       \\
S3DIS      & 4,096 &    KNN   &       6  &     3   & 45\%/45\%/10\% &  Point Cloud Network   \\ \midrule[0.3pt]
\end{tabular}
}}
\vspace{-0.2cm}
\end{table*}

    \begin{table*}[htbp]
    \centering
    \caption{A summary of our proposed graph-specfic data simulation strategies.}
    \resizebox{\linewidth}{20mm}{
\setlength{\tabcolsep}{4mm}{
\begin{tabular}{cccccc}
\midrule[0.3pt]
Federated Simulation                & Scenarios   & Feature & Label & Topology & Implemented By\\ \midrule[0.3pt]
Feature Distribution Skew           & Both        & \textcolor{teal}{\usym{2713}}             & \textcolor{red}{\usym{2717}}           & \textcolor{red}{\usym{2717}}              & FS-G, OpenFGL \\
Label Distribution Skew             & Both        & \textcolor{red}{\usym{2717}}             & \textcolor{teal}{\usym{2713}}           & \textcolor{red}{\usym{2717}}             & FS-G, FedGraphNN, OpenFGL \\
Cross Domain Data Skew              & Both        & \textcolor{teal}{\usym{2713}}             & \textcolor{teal}{\usym{2713}}           & \textcolor{red}{\usym{2717}}             & FS-G, OpenFGL \\ \midrule[0.3pt]
Topology Distribution Skew          & Graph-FL    & \textcolor{red}{\usym{2717}}             & \textcolor{red}{\usym{2717}}           & \textcolor{teal}{\usym{2713}}             & OpenFGL \\
Metis-based Community Split         & Subgraph-FL & \textcolor{red}{\usym{2717}}             & \textcolor{red}{\usym{2717}}           & \textcolor{teal}{\usym{2713}}             & FS-G, OpenFGL \\
Louvain-based Community Split       & Subgraph-FL & \textcolor{red}{\usym{2717}}             & \textcolor{red}{\usym{2717}}           & \textcolor{teal}{\usym{2713}}             & FS-G, OpenFGL \\
Metis-based Label Imbalance Split   & Subgraph-FL & \textcolor{red}{\usym{2717}}             & \textcolor{teal}{\usym{2713}}           & \textcolor{teal}{\usym{2713}}             & OpenFGL \\
Louvain-based Label Imbalance Split & Subgraph-FL & \textcolor{red}{\usym{2717}}             & \textcolor{teal}{\usym{2713}}           & \textcolor{teal}{\usym{2713}}             & OpenFGL \\ \midrule[0.3pt]
\end{tabular}
    \label{tab: simulation}}}
    \vspace{-0.2cm}
    \end{table*}
    
\textbf{Simulation Strategies.} 
    In response to policy constraints on acquiring distributed graphs, we draw inspiration from federated learning in computer vision to simulate generalized federated scenarios by partitioning existing datasets into distributed subsets.
    This strategy is similar to recent FGL benchmarks~\cite{he2021fedgraphnn,WangFedScope_22_fsg}.
    In our proposed OpenFGL, we integrate 8 federated data simulation strategies driven by practical applications, in which we ensure the graph data distributed to each client exhibits similar patterns in feature, label, or topology while maintaining a controllable heterogeneity across clients. 
    Specifically, for both Graph-FL and Subgraph-FL, we implement 3 simulation strategies widely used in graph-independent FL.
    In the context of Graph-FL, we introduce a topology-oriented simulation strategy called Topology Shift, which distributes graphs based on degree distribution. 
    The inspiration for this approach stems from key insights offered by recent FGL studies~\cite{xie2021gcfl, tan2023fedstar}: in Graph-FL, the structure Non-iid resulting from topology shift is the primary challenge in collaborative optimization, where node features and labels appear less significant by comparison.
    As for Subgraph-FL, existing strategies predominantly utilize community detection algorithms such as Louvain~\cite{blondel2008louvain} and Metis~\cite{karypis1998metis} to identify clusters with dense intra-community connections. 
    Then, these clusters' nodes are subsequently allocated to local clients to construct corresponding induced subgraphs for partitioning.
    These strategies all operate under a common assumption: that the private data collected by each local agent in the real world contains dense internal connections but is loosely connected across clients~\cite{hsiao1992_federated_database_2,kamel1992_federated_database_3,li2020_federated_database_5,li2021_federated_database_7}.
    Despite their effectiveness, limitations persist in their application. 
    This is because Subgraph-FL primarily focuses on node-level and link-level tasks, where node profiles (i.e., node features and labels) are crucial. 
    However, these strategies do not consider label distribution for federated data simulation.
    Hence, we introduce Metis-based Label Imbalance Split and Louvain-based Label Imbalance Split. These methods, refining the aforementioned strategies, carefully consider label distribution during cluster allocation to better simulate realistic and generalized distributed scenarios. We outline these strategies in Table~\ref{tab: simulation}, with descriptions provided below:

    \textbf{Feature Distribution Skew} is a graph-independent strategy to simulate feature distribution shifts~\cite{wang2022fl_cite11,jiang2024fl_cite12}. 
    In OpenFGL, we utilize this approach to create more challenging and realistic scenarios for evaluating FGL algorithms~\cite{li2022non-iid_fl_survey3}.
    Specifically, we implement various feature operations:
    (1) Adding Gaussian or Laplacian noise to introduce variability;
    (2) Applying scaling operations to simulate different magnitudes of features;
    (3) Employing mathematical transformations to further diversify the feature distributions.

    \textbf{Label Distribution Skew}
    is a graph-independent strategy~\cite{wu2024fl_cite9,yao2023fl_cite7}. 
    In our implementation, we use the $\alpha$-based Dirichlet distribution to create imbalanced label distributions across clients~\cite{blei2003LDA}. 
    This approach ensures varied and imbalanced label distributions, simulating real-world data scenarios. 
    The $\alpha$ in Dirichlet distribution controls the concentration of probabilities across label classes, with larger values leading to more uniform distributions and smaller values resulting in sparser and more concentrated distributions.

    \textbf{Cross Domain Data Skew} is a fundamental challenge in FL, arising from the heterogeneous nature of data sources and collection methods across distributed databases~\cite{yang2019_federated_database_4}. 
    In OpenFGL, we simulate this scenario by evenly distributing multiple datasets among a predefined number of clients, maintaining a diverse representation.

    \textbf{Topology Distribution Skew} represents the strategy for partitioning graphs based on their topology properties in Graph-FL~\cite{xie2021gcfl,tan2023fedstar}.
    This approach involves sorting the global graph according to specific characteristics, such as average node degree, and then distributing the resulting graphs to predefined clients. 
    This ensures that the distribution of graph data among clients accurately reflects the underlying topological diversity of the original global dataset.

    \textbf{Metis-based Community Split} is a widely adopted Subgraph-FL federated data simulation strategy that utilizes a multilevel recursive bisection and k-way partitioning technique. 
    This method iteratively reduces the size of the graph and refines the partitioning.
    Notably, compared to the following the Louvain-based data simulation strategy, Metis can directly partition a graph into a predefined number of communities, aligning precisely with the number of clients and streamlining data allocation in the federated settings.

    \textbf{Louvain-based Community Split} stands as the other prevalent federated data simulation strategy in Subgraph-FL, partitioning a graph into multiple communities (subgraphs) via modularity optimization. 
    The number of communities is determined by the resolution parameter of the Louvain algorithm.
    However, the Louvain algorithm often generates more communities than the predefined number of clients. To resolve this, communities can be allocated among the clients by averaging node quantities.

    \textbf{Metis-based Label Imbalance Split} is a new Subgraph-FL data simulation strategy introduced in this paper. 
    The naive Metis-based Community Split lacks post-processing capabilities, leading to challenges in controlling subgraph heterogeneity among clients. 
    In contrast, our approach enables predefined community partitioning, followed by clustering based on label distribution similarity, thereby consolidating similar communities under a single client.

    \textbf{Louvain-based Label Imbalance Split} enhances the conventional Louvain method by allocating communities to clients based on similarities in label distributions rather than solely on node averages. 
    This approach ensures that each client receives communities with consistent label characteristics, thereby mitigating label imbalance and promoting equitable model training across federated clients. 
    By aligning label distributions, this strategy enhances the fairness and robustness of federated learning, reducing biases that may arise from heterogeneous label distributions among clients.

    These 8 federated distributed data simulation strategies, meticulously developed based on the combination of features, labels, and topology, significantly enhance the robustness and generalization capabilities of FGL studies.
    We have crafted a comprehensive and realistic benchmark tailored for industrial applications, thereby fostering substantial progress and paving the way for future advancements in FGL research. 
    Therefore, OpenFGL not only tests the effectiveness of existing FGL algorithms but also serves as a platform for developing new methodologies and approaches.
    
    \textbf{Downstream Tasks.}
    Our proposed OpenFGL evaluates FGL studies across a range of downstream tasks, including graph classification and regression for Graph-FL, as well as node classification, clustering, and link prediction for Subgraph-FL. While acknowledging the traditional focus on graph and node classification, OpenFGL extends its scope to additional tasks, promoting broader advancements and greater flexibility in the FGL benchmark.
    Notably, to avoid complex presentation and ensure reader-friendly, we primarily focus on classification task to present experimental results.

\subsection{Method-level FGL Algorithms}
\label{sec: Method-level FGL Algorithms}
    {\textbf{GNN Backbones.}} 
    To provide a broader spectrum of learning paradigms on the client side, OpenFGL integrates a diverse range of local GNN backbones.
    Specifically, we implement various well-designed polling strategies (TopKPooling~\cite{gao2019_TopKPooling}, SAGPooling~\cite{lee2019_SAGPooling}, EdgePooling~\cite{diehl2019_EdgePooling}, and PANPooling~\cite{ma2020_PANPooling}) based on the most representative GIN~\cite{xu2018gin} with weight-free MeanPooling~\cite{xu2018gin} for Graph-FL.
    As for Subgraph-FL, OpenFGL includes prevalent GCN~\cite{kipf2016gcn}, GAT~\cite{velivckovic2017gat}, GraphSAGE~\cite{hamilton2017graphsage}, SGC~\cite{wu2019sgc} and GCNII~\cite{chen2020gcnii}. 
    The detailed descriptions of these backbones can be found in ~\cite{OpenFGL} (\ref{appendix: Baseline Description}). 

    \noindent{\textbf{FL/FGL Algorithms.}} 
    To achieve federated training, multi-client collaboration algorithms are crucial (CV-based FL also can be applied to FGL). 
    We follow the historical progression of FL to include a spectrum of algorithms from the most representative methods in CV to FGL:
    (1) {CV-based FL algorithms}: FedAvg~\cite{mcmahan2017fedavg}, FedProx~\cite{li2020fedprox}, Scaffold~\cite{karimireddy2020scaffold}, MOON~\cite{li2021moon}, FedDC~\cite{gao2022feddc}, FedProto~\cite{tan2022fedproto}, FedNH~\cite{dai2023fednh} and FedTGP~\cite{zhang2024fedtgp};  
    (2) {all FGL algorithms possible}: GCFL+~\cite{xie2021gcfl} and FedStar~\cite{tan2023fedstar} in Graph-FL and FedSage+~\cite{zhang2021fedsage}, Fed-PUB~\cite{baek2022fedpub}, FedGTA~\cite{li2023fedgta}, FGSSL~\cite{huang2023fgssl}, FedGL~\cite{chen2024fedgl}, AdaFGL~\cite{li2024adafgl}, FGGP~\cite{wan2024fggp}, FedDEP~\cite{zhang2024feddep}, and FedTAD~\cite{zhu2024fedtad} in Subgraph-FL.
    More details about these algorithms can be found in ~\cite{OpenFGL} (\ref{appendix: Baseline Description}).
    Notably, these algorithms are implemented with a unified API to facilitate future development in a user-friendly manner.
    For more details about our API design from the algorithm perspective, please refer to Sec.~\ref{sec: OpenFGL Algorithm Tutorial}.

\subsection{Experiment-level FGL Evaluations}
\label{sec: Experiment-level FGL Evaluations}
    \textbf{Data Analysis.} 
    (1) Feature KL Divergence: 
    It reveals feature skew among clients while the label domain remains consistent, which may arise due to the different geographical locations of clients, such as the characteristics of a certain disease may significantly differ across various regions.
    (2) Label Distribution:
    It is widely discussed in CV-based FL. 
    However, in FGL, the relationship between labels and topology frequently reveals underlying connections, characterized as homophily.
    Thus, we further integrate multi-level homophily metrics~\cite{pei2020geomgcn, 2021linkx, platonov2023hete_gnn_survey4, platonov2022hete_gnn_survey5, luan2022hete_gnn_survey2} to offer comprehensive analysis.
    (3) Topology Statics:
    This perspective stems from the critical role of topology in GNNs, especially in distributed scenarios. 
    This arises from the significant impact of diverse local topology statistics on the local model, causing complex cascading effects in collaboration. Therefore, we examine topological differences among clients (e.g., Degree, Centrality, Largest Component) to provide insights.

    % \textcolor{magenta}{\textbf{effectiveness}}, \textcolor{teal}{\textbf{robustness}}, and \textcolor{orange}{\textbf{efficiency}}. 

    \noindent
    \textbf{Effectiveness.} 
    Details of our evaluation metrics are as follows: graph/node classification (Accuracy, F1, Recall, Precision), graph regression (MSE, RMSE), link prediction (AP, AUC-ROC), and node clustering (Clustering-accuracy, NMI, ARI).
    More detailed descriptions of these metrics are presented in ~\cite{OpenFGL} (\ref{appendix: Metric Description}). 

    \noindent
    \textbf{Robustness.} 
    To evaluate the practical deployment of FGL, we examine its robustness from the following perspectives: 
    (1) \textbf{Noise}: 
    This corresponds to data quality issues resulting from data collection~\cite{li2020_federated_database_5, liu2022_federated_database_9}.
    (2) \textbf{Sparsity}:
    This reflects scenarios with incomplete features, labels, and topology due to data scarcity and high labor costs, and with a low rate of client participation~\cite{gu2023fl_cite5, wu2024fl_cite9, li2024fl_cite2}.
    (3) \textbf{Client Communication}:
    This simulates scenarios with network constraints or high communication costs~\cite{yao2023fl_cite7,yuan2024fl_cite3}.
    (4) \textbf{Generalization}: 
    This evaluates the effectiveness of algorithms in various scenarios.
    (5) \textbf{Privacy Preserve}: 
    This reflects the applicability of algorithms in privacy-sensitive scenarios, thereby we conduct an in-depth analysis from the perspective of Differential Privacy (DP).
    Please refer to ~\cite{OpenFGL} (\ref{appendix: Robustness Simulation Description}) for further details on the robustness settings.

    \noindent
    \textbf{Efficiency.}
    To facilitate FGL deployment, we conduct an evaluation of current baselines regarding their efficiency.
    Specifically, we evaluate them from both theoretical (algorithm complexity) and experimental (communication cost and running time) perspectives.

\section{Experiments and Analysis}
    In this section, we systematically investigate FGL algorithms by answering the following questions:
    (1) For \textbf{Effectiveness}, 
    \textbf{Q1}: What advantages does federated collaboration offer compared to training solely on local data?
    \textbf{Q2}: How do FGL algorithms and federated implementations of GNNs perform in two prevalent FGL scenarios?
    (2) For \textbf{Robustness},
    \textbf{Q3}: How do FGL algorithms perform under local noise and sparsity settings (i.e., features, labels, and edges)?
    \textbf{Q4}: What are the performance of FGL algorithms under low client participation rates in communication?
    \textbf{Q5}: Can FGL algorithms maintain generalization across various graph-specific distributed scenarios?
    \textbf{Q6}: Do FGL algorithms support additional DP privacy protection?
    (3) For \textbf{Efficiency},
    \textbf{Q7}: What are the theoretical algorithm complexity of FGL algorithms?
    \textbf{Q8}: What is the practical running efficiency of FGL algorithms?
    To maximize the usage for the constraint space, more results are shown in~\cite{OpenFGL} (\ref{appendix: Effectiveness Evaluation Strategies}-\ref{appendix: Hyperparameter Settings}). 

\subsection{Performance Comparison}
\label{sec: Performance Comparison}
    To answer \textbf{Q1}, in addition to the federated multi-client collaboration, we introduce "Local" to represent solely local training for analyzing the advantages and potential limitations of FGL. 
    Based on this, to answer \textbf{Q2}, we present the end-to-end performance in Table~\ref{tab: graph-fl end-to-end comparison} and Table~\ref{tab: subgraph-fl end-to-end comparison}. 
    The detailed analysis is presented as follows:

\noindent
\textbf{Graph-FL Scenario}.
    Since the baselines for Graph-FL are scarce. 
    Therefore, we expand pooling-based backbones and CV-based FL in Table~\ref{tab: graph-fl end-to-end comparison}. 
    For \textbf{Q1}, we find that the benefits of federated collaboration are not significant for small-scale MUTAG, BZR, and COX2 due to limited data that can not support federated training and thus affect predictions. 
    Subsequently, we conclude that \textit{C1: Federated collaboration is more advantageous for larger-scale datasets, benefiting from abundant data sources~\cite{ruan2024conclusion_1_1,li2024conclusion_1_2}}.
    As for \textbf{Q2}, we observe that: 
    (1) GCFL+ and FedStar concentrate on topology Non-iid within the cross-domain simulation, thereby not consistently achieving competitive performance in this single-source setting.
    Therefore, they perform less favorably than FL algorithms on ENZYMES, COLLAB, and MULTI.
    (2) Existing FGL algorithms heavily rely on node semantics. 
    We observe significant improvements on datasets with abundant node descriptions like DD and PROTEINS, whereas limited performance on BINARY whose node representations are initialized with node degrees.
    Consequently, we conclude that \textit{C2: Graph-FL algorithms still have improvement space, especially in the single-source domain and constrained data semantics~\cite{abushofa2023conclusion_2_1,li2024conclusion_2_2}.}

\begin{table*}[t]
\caption{Graph-FL test accuracy (\%).
The best result is \textbf{bold}.
The second result is \underline{underlined}.}
\label{tab: graph-fl end-to-end comparison}
\resizebox{\linewidth}{36mm}{
\setlength{\tabcolsep}{2mm}{
\begin{tabular}{cccccccccc}
\midrule[0.3pt]
Graph-FL   & MUTAG                 & BZR                   & COX2                  & ENZYMES               & DD                    & PROTEINS              & COLLAB                & BINARY                & MULTI    \\ \midrule[0.3pt]
GIN-Local  & 84.2±2.5              & 84.3±1.6              & 81.6±2.9              & 40.7±1.1              & 82.7±2.0              & 81.8±1.1              & 75.4±1.9              & 76.3±2.8              & 47.1±1.8 \\
SAG-Local  & 82.0±1.9              & \textbf{89.5±1.9}     & \underline{82.1±2.3}  & 41.4±0.8              & 80.3±1.5              & 83.3±1.4              & 77.0±1.5              & 77.7±2.5              & 49.0±2.1 \\
Edge-Local & 80.8±1.9              & 86.7±1.8              & 78.6±2.4              & 42.1±0.7              & 81.5±2.4              & 82.7±0.9              & 76.2±1.7              & 78.9±1.9              & 48.0±2.0 \\
PAN-Local  & \textbf{86.1±2.7}     & 80.3±2.0              & 80.8±2.4              & 38.0±1.0              & 84.2±2.2              & 81.0±1.9              & 75.1±2.0              & 80.7±2.3              & 46.1±2.4 \\ \midrule[0.3pt]
FedAvg     & 78.9±2.9              & 76.5±1.3              & 79.0±1.7              & 47.4±0.9              & 82.4±2.6              & 80.1±1.5              & 77.5±1.6              & 79.2±2.5              & 50.8±2.0 \\
FedProx    & 76.5±2.4              & 81.8±1.7              & 77.2±1.6              & 46.7±1.4              & 83.1±1.5              & 77.4±1.7              & 77.9±1.9              & \underline{81.9±2.0}  & 51.6±2.2 \\
Scaffold   & 75.4±2.9              & 82.3±1.8              & 82.0±1.4              & 40.9±1.5              & 84.5±2.4              & 79.9±1.1              & 76.4±1.8              & 80.3±1.7              & \textbf{52.4±2.8} \\
MOON       & 80.5±2.9              & 82.6±1.8              & 78.8±1.5              & \textbf{49.3±1.6}     & 79.8±2.1              & 80.0±2.0              & \textbf{79.8±2.0}     & 81.1±2.0              & 51.4±2.2 \\
FedProto   & 82.7±2.0              & 86.7±1.4              & 79.4±2.4              & 42.5±1.4              & 85.2±2.0              & 80.3±1.3              & 76.7±1.4              & 80.6±2.7              & 49.9±2.1 \\
FedNH      & 84.3±2.2              & 85.2±1.6              & 81.2±2.4              & 45.3±1.5              & 84.9±2.0              & 81.2±1.8              & 75.3±1.2              & 79.4±2.0              & 50.4±2.6 \\
FedTGP     & 83.8±2.8              & 84.6±1.0              & 81.8±1.6              & 43.0±1.0              & \underline{87.3±2.8}  & 80.9±2.0              & 77.2±2.1              & 78.6±2.5              & 50.9±2.5 \\ \midrule[0.3pt]
GCFL+      & 82.6±2.6              & 87.8±1.9              & \textbf{82.6±2.3}     & 47.8±1.3              & 85.2±2.5              & \underline{83.6±1.3}  & 77.5±1.1              & 80.4±2.3              & \underline{51.8±2.5} \\
FedStar    & \underline{84.7±2.6}  & \underline{89.1±1.5}  & 80.6±2.3              & \underline{48.4±0.8}  & \textbf{88.4±2.3 }    & \textbf{84.5±1.7}     & \underline{78.6±1.7}  & \textbf{82.7±2.3}     & 51.4±2.6 \\ \midrule[0.3pt]
\end{tabular}
}}
\end{table*}

\begin{table*}[t]
\caption{Subgraph-FL test accuracy (\%).
The best result is \textbf{bold}.
The second result is \underline{underlined}.}
\label{tab: subgraph-fl end-to-end comparison}
\resizebox{\linewidth}{36mm}{
\setlength{\tabcolsep}{2mm}{ 
\begin{tabular}{cccccccccc}
\midrule[0.3pt]
Subgraph-FL & Cora                      & CiteSeer                  & PubMed                     & Photo                       & Computers                   & Products                & Chameleon              & Actor                 & Ratings  \\ \midrule[0.3pt]
GCN-Local   & 77.9±0.3                  & 64.3±0.8                  & 84.6±0.3                   & 88.8±0.4                    & 87.7±0.6                    & 79.4±0.5                & 64.7±0.6               & 29.4±1.4              & 45.5±0.8 \\
GAT-Local   & 78.5±0.4                  & 63.9±0.6                  & 85.3±0.4                   & 89.6±0.5                    & 87.4±0.5                    & 78.9±0.6                & 65.1±0.5               & 30.2±0.9              & \underline{46.2±0.5} \\ \midrule[0.3pt]
FedAvg      & 82.5±0.7                  & 69.5±0.7                  & 86.4±0.5                   & 90.3±0.7                    & 89.1±0.4                    & 82.3±0.5                & 61.2±0.8               & 28.7±0.8              & 42.5±0.4 \\
FedDC       & 81.4±0.8                  & 70.4±0.5                  & 87.9±0.4                   & 91.2±0.6                    & 88.4±0.5                    & 81.9±0.3                & 58.6±1.2               & 26.9±1.2              & 41.2±0.4 \\
FedProto    & 79.4±0.6                  & 67.2±0.2                  & 85.1±0.2                   & 87.4±0.4                    & 86.9±0.3                    & 77.2±0.4                & 64.0±0.6               & 28.0±0.6              & 46.1±0.4 \\
FedTGP      & 80.7±0.5                  & 68.8±0.4                  & 85.9±0.3                   & 86.5±0.5                    & 86.4±0.6                    & 78.3±0.5                & 62.7±1.1               & 28.4±0.7              & 45.7±0.9 \\ \midrule[0.3pt]
FedSage+    & 82.6±0.8                  & 71.2±0.8                  & \textbf{88.2±0.7}          & 90.8±0.8                    & 90.4±0.8                    & \underline{82.8±0.7}    & 65.6±0.7               & 30.8±1.0              & 45.8±0.7 \\
FedGTA      & 83.0±0.4                  & \textbf{72.4±0.4}         & 87.6±0.4                   & 91.0±0.4                    & 90.8±0.5                    & \textbf{83.2±0.4}       & \underline{66.2±0.8}   & 30.5±0.6              & 45.5±0.5 \\
Fed-PUB     & 81.7±0.6                  & 71.9±0.7                  & 87.8±0.3                   & \textbf{91.5±0.6}           & \underline{91.1±0.7}        & 82.1±0.5                & 64.4±0.7               & 29.2±0.8              & 44.8±0.6 \\
FGSSL       & 81.5±0.8                  & 70.1±0.5                  & 87.3±0.4                   & 88.8±0.6                    & 89.2±0.5                    & OOM                     & 64.9±0.9               & 28.9±1.0              & 45.2±0.8 \\
FedGL       & 82.5±0.7                  & 71.5±0.7                  & 87.1±0.5                   & 89.7±0.5                    & 90.7±0.5                    & OOM                     & 65.4±0.8               & 29.4±1.3              & 46.0±0.7 \\
AdaFGL      & \underline{83.4±0.5}      & \underline{72.0±0.5}      & 87.9±0.6                   & 91.3±0.7                    & 91.0±0.6                    & OOM                     & \textbf{67.2±1.0}      & \textbf{31.2±1.2}     & \textbf{46.9±0.5} \\
FGGP        & 81.4±0.5                  & 69.1±0.7                  & 87.0±0.4                   & 88.6±0.6                    & 88.3±0.4                    & OOM                     & 65.3±0.7               & 28.3±0.4              & 45.6±0.5 \\
FedTAD      & \textbf{84.1±0.6}         & 71.8±0.8                  & \underline{88.0±0.6}       & \underline{91.4±0.5}        & \textbf{91.6±0.7}           & 81.9±0.3                & 65.9±1.3               & \underline{30.7±1.0}  & 46.1±0.8 \\ \midrule[0.3pt]
\end{tabular}
}}
\vspace{0.25cm}
\end{table*}

\noindent
\textbf{Subgraph-FL Scenario}.
    Experimental results are shown in Table~\ref{tab: subgraph-fl end-to-end comparison}. 
    For \textbf{Q1}, although Subgraph-FL can benefit from abundant data, for Chameleon, Actor, and Ratings, the improvement from both FL and FGL is limited or even worse than solely local training in some cases. 
    We attribute this to the heterophily, where differences in node connection rules across clients significantly affect local updates and server-side collaboration, deviating from the global optimum and resulting in sub-optimal performance. 
    Although AdaFGL addresses this issue through personalized technology, there is still room for better performance. 
    Based on this and \textit{C1}, we conclude that \textit{C3: The prerequisite for positive impacts of federated collaboration is the uniform distribution of node features, labels, and topology~\cite{liu2024conclusion_3_1,xia2024conclusion_3_2}}. 
    Regarding \textbf{Q2}, we observe that: 
    (1) Subgraph-FL is thriving, with numerous baselines vigorously competing for the best performance. 
    Among them, FedTAD and AdaFGL stand out in most cases. 
    (2) The outstanding performance of existing algorithms stems from the fine-grained exploration of the topology, but some methods lack scalability when dealing with large-scale ogbn-products, resulting in OOM (out-of-memory) errors. 
    Consequently, we conclude that \textit{C4: Subgraph-FL algorithms need to resolve the complexity in real-world deployments, especially focusing on large-scale scenarios and graph-specific federated heterogeneity challenges~\cite{li2023conclusion_4_1,li2024conclusion_4_2}}.

\subsection{Robustness Analysis}
\label{sec: Robustness Analysis}

\begin{figure*}[t]
	\centering
    \setlength{\abovecaptionskip}{0.3cm}
  \includegraphics[width=\textwidth]{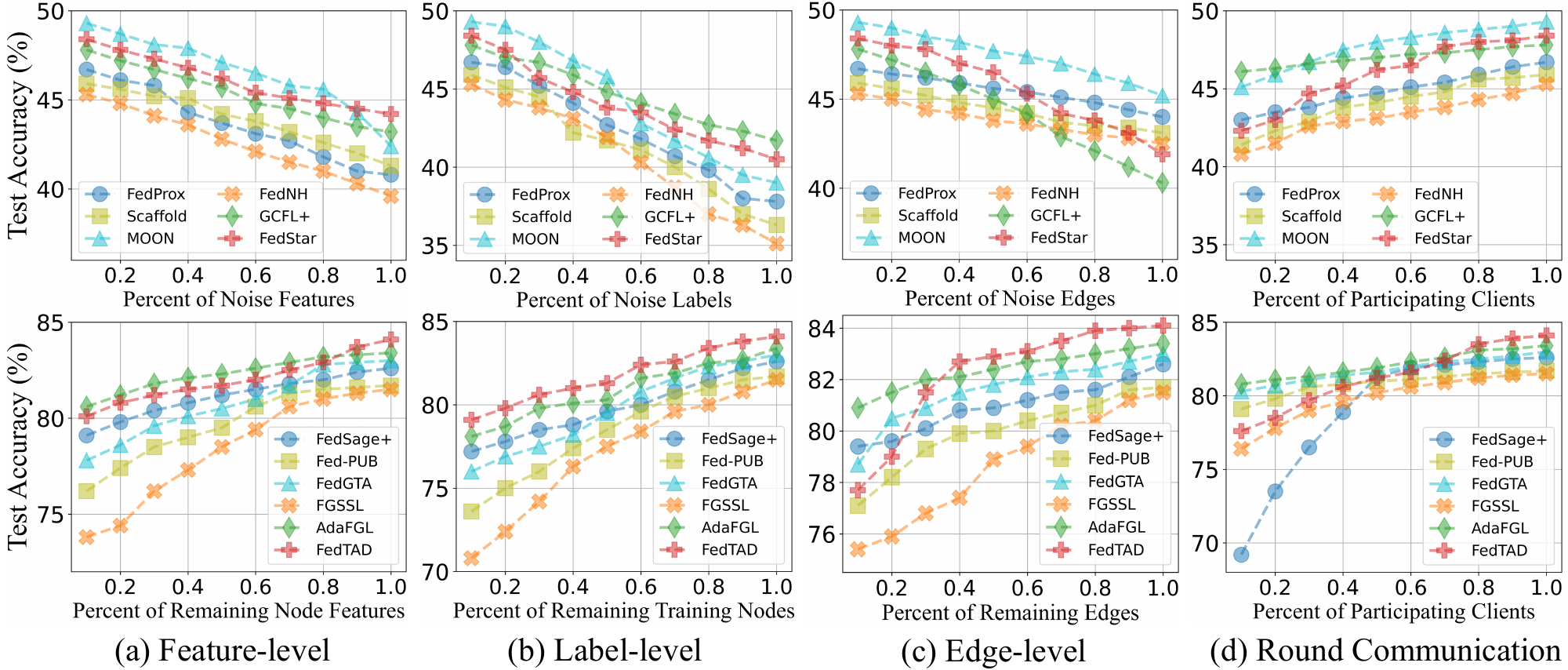}
  \caption{
    Robustness performance on Graph-FL ENZYMES (upper) and Subgraph-FL Cora (lower).
}
  \label{fig: exp_noise_sparsity_client}
\end{figure*}

    % To concisely answer \textbf{Q3} and \textbf{Q4} and ensure reader-friendly presentation without complex charts, we provide in Fig.~\ref{fig: exp_noise_sparsity_client} the impact of local noise settings in Graph-FL and local sparsity settings in Subgraph-FL, as well as the effect of variations in the proportion of participating clients in each communication round in both scenarios on baseline prediction performance.

\begin{table*}[t]
\caption{Generalization performance (\%).
The best result is \textbf{bold}.
The second result is \underline{underlined}.}
\label{tab: data simulation generalization}
\resizebox{\linewidth}{38.5mm}{
\setlength{\tabcolsep}{0.9mm}{
\begin{tabular}{c|ccc|ccc|ccc|c}
\midrule[0.3pt]
Graph-FL    & \multicolumn{3}{c|}{DHFR}                  & \multicolumn{3}{c|}{DD}                    & \multicolumn{3}{c|}{COLLAB}                & Multi-source \\
Simulation  & Feature                   & Label (0.5)               & Topology                  & Feature           & Label (0.5)           & Topology  & Feature  & Label (0.5) & Topology & Cross-domain   \\ \midrule[0.3pt]
FedAvg      & 69.1±2.7                  & 70.5±3.6                  & 72.3±3.2                  & 80.3±2.9          & 81.2±2.4              & {81.8±2.4}  & 68.7±2.5          & 76.6±1.8             & 70.3±1.8     & 73.6±2.6       \\
FedDC       & 70.3±3.0                  & 68.2±4.2                  & 71.6±3.5                  & 78.6±4.2          & 83.9±3.2              & 80.7±3.0           & 66.9±3.9          & \underline{78.2±2.2}   & 69.8±2.7     & 71.3±3.2       \\
FedProto    & 68.4±2.4                  & 71.9±3.1                  & 71.1±3.2                  & 79.8±2.5          & 83.6±2.5              & 82.1±2.5           & 70.4±1.5          & 76.0±1.8             & 72.0±1.1     & 72.6±2.8       \\
FedTGP      & 71.5±2.7                  & 69.8±3.5                  & 72.0±2.9                  & 81.4±2.9          & \underline{85.5±2.1}  & 80.8±3.6           & 72.9±2.6          & 75.9±2.3             & 72.7±1.9     & 74.1±3.0       \\
GCFL+       & \underline{72.3±1.9}      & \underline{73.2±2.4}      & \textbf{74.5±2.4}         & \textbf{83.1±2.2} & 84.2±1.8              & \textbf{83.9±2.7}    & \underline{73.8±1.9} & {77.3±2.5}    & \textbf{75.4±1.5}     & \underline{76.2±2.5}       \\
FedStar     & \textbf{73.9±2.6}         & \textbf{74.6±2.9}         & \underline{73.4±3.1}      & \underline{82.5±2.9} & \textbf{86.3±2.0}  & \underline{82.2±3.2}  & \textbf{74.6±2.3} & \textbf{78.5±2.9}    & \underline{73.8±2.6}     & \textbf{77.3±2.9}       \\ \midrule[0.3pt]
Subgraph-FL & \multicolumn{3}{c|}{Photo}                 & \multicolumn{3}{c|}{Squirrel}              & \multicolumn{3}{c|}{Questions}             & Multi-source \\
Simulation  & Louvain                   & Metis+               & Louvain+           & Louvain           & Metis+               & Louvain+           & Louvain           & Metis+                        & Louvain+                   & Cross-domain   \\ \midrule[0.3pt]
FedGL       & 91.4±1.0                  & 89.5±0.9             & 87.9±1.1           & 47.0±1.5          & \underline{47.2±1.0} & 45.7±1.3           & 96.8±0.4          & 94.4±0.4                          & \textbf{93.5±0.3}          & 78.3±0.8       \\
FedSage+    & \underline{92.1±0.7}      & 90.4±0.8             & \textbf{88.9±1.0}  & 46.2±1.4          & 46.0±1.2             & \textbf{46.4±1.1}  & \underline{97.0±0.3}  & 93.6±0.5                      & 92.4±0.4                   & 80.5±0.6       \\
FGSSL       & 90.5±0.6                  & 89.8±0.6             & 87.5±0.7           & 45.9±0.7          & 46.3±0.9             & 43.3±0.8           & 96.1±0.4          & 94.2±0.4                          & 92.7±0.3                   & 76.9±1.0       \\
FedGTA      & \textbf{92.4±0.4}         & \textbf{91.0±0.5}    & \underline{88.6±0.4} & \underline{47.8±0.8}  & 47.0±0.5        & 45.8±0.6           & 96.3±0.2          & \textbf{95.0±0.3}                & \underline{95.3±0.2}        & \textbf{82.4±0.4}       \\
Fed-PUB     & 91.8±0.5                  & 90.2±0.7             & 88.3±0.6           & 47.5±1.0          & 46.9±1.3             & 44.6±1.1           & 96.5±0.3          & 94.1±0.4                          & 95.0±0.2                   & 79.8±0.7       \\
AdaFGL      & 91.7±0.8                  & \underline{90.9±0.7} & 88.2±0.8           & \textbf{48.6±1.2}   & \textbf{47.8±1.5}   & \underline{46.0±0.9} & \textbf{97.2±0.2}   & \underline{94.7±0.3}         & 95.2±0.3              & \underline{81.6±0.9}       \\ \midrule[0.3pt]
\end{tabular}
}}
\end{table*}

\textbf{Local Noise}.
    To answer \textbf{Q3} from the perspective of noise, the experimental results are shown in the upper part of Fig.~\ref{fig: exp_noise_sparsity_client}(a)-(c). 
    For feature noise, we randomly select the channels of node features and inject Gaussian noise. 
    To simulate edge noise in the Graph-FL without node labels, we utilize Metattack~\cite{zugner_adversarial_2019_metaattck} to add noise edges, significantly perturbing the training gradients. 
    As for label noise, we randomly reassign non-real labels to a certain proportion of training node samples.
    Based on the experimental results, we observe that FGL algorithms are highly sensitive to edge noise compared to topology-agnostic FL algorithms. 
    This inherent limitation directly disrupts the model optimization of GCFL+ and FedStar, misleading local updates and topology-driven collaboration, thereby resulting in sub-optimal predictive performance. 
    However, GCFL+ and FedStar demonstrate superior robustness under feature and label noise, as they address client interference with each other using server-side clustering customization and additional local models maintained at client-side.
    Consequently, we can conclude that \textit{C5: Noise scenarios determine the performance lower bound for FGL algorithms, where personalized strategies emerge as crucial technologies. 
    However, they fall slightly short in addressing edge noise~\cite{zhang2024conclusion_5_1,lin2024conclusion_5_2}.}

\vspace{0.1cm}
\noindent
\textbf{Local Sparsity}.
    To answer \textbf{Q3} from the perspective of sparsity, the experimental results are shown in the lower part of Fig.~\ref{fig: exp_noise_sparsity_client}(a)-(c).
    Regarding feature sparsity, we simulate partial feature absence for unlabeled nodes and randomly remove the original edges.
    As for label sparsity, we change the ratio of training nodes. 
    Under feature sparsity, FGSSL and Fed-PUB exhibit significant performance fluctuations due to heavy reliance on high-quality features for node-wise contrastive learning and model-wise gradient matching. 
    Conversely, FedSage+, AdaFGL, and FedTAD leverage multi-client collaboration, mitigating confusion in under-trained model collaboration, thus ensuring robustness.
    Similar analysis can extend to label sparsity, as both of them directly affect local updates.
    Regarding edge sparsity, FedTAD's vulnerability lies in its reliance on client-side topology embeddings to supervise server-sider pseudo-subgraph generator.
    However, AdaFGL and FedSage+, leveraging topology mining, adeptly handle this challenge.
    Based on this, we can conclude that: \textit{C6: Sparsity scenarios determine the performance upper bound for FGL algorithms, where multi-client collaboration is the pivotal technology, particularly in synergy with topology mining~\cite{wu2024conclusion_6_1,babakniya2023conclusion_6_2}.}

\noindent
\textbf{Client Participation}.
    To answer \textbf{Q4}, we present the experimental results in Fig.~\ref{fig: exp_noise_sparsity_client}(d), where robust FGL algorithms with low client participation exhibit one of the following characteristics: 
    (1) They rely less on messages received from the server and focus on local training; 
    (2) They custom global messages for each participating client. 
    For instance, in Graph-FL, the unstable performance of FedStar arises from its heavy reliance on specific topological properties. 
    Conversely, GCFL+ ensures high-quality local updates by tailoring the most suitable messages for each client through server-side clustering.
    In Subgraph-FL, AdaFGL, Fed-PUB, and FedGTA rely on client-side personalized training, server-side pseudo-graph-driven clustering, or identification of subgraph statistics to ensure custom global messages for each participating client. 
    Consequently, we can conclude that \textit{C7: Low client participation underscores the emphasis of FGL algorithms on local updates, highlighting the importance of local data understanding and customizing messages for each client~\cite{fu2023conclusion_7_1,qu2023conclusion_7_2}.}

\noindent
\textbf{Generalization}.
    To answer \textbf{Q5}, the experimental results are shown in Table~\ref{tab: data simulation generalization}, where graph-level cross-domain setting includes MUTAG, COX2, PTC-MR, AIDS, ENZYMES, DD, PROTEINS, COLLAB, and IMDB-B/M and subgraph-level setting includes Cora, CiteSeer, Pubmed, CS, and Physics.
    We observe that current FGL algorithms exhibit inconsistent performance across data simulations. 
    Specifically, in Graph-FL, tiny-scale datasets with abundant node descriptions such as DHFR and DD mitigate inherent differences among data simulations potentially due to over-fitting issues. 
    %Notably, this result may also stem from over-fitting issues, causing FGL algorithms to inadvertently overlook subtle differences between different data simulations.%
    As for the Subgraph-FL, data simulations present challenges for existing FGL algorithms. 
    For instance, Fed-PUB and AdaFGL, incorporating personalized strategies, lose their previous advantages, whereas FedSage+ and FedGTA, emphasizing multi-client collaboration mechanisms, show significant potential.
    Therefore, we can conclude that \textit{C8: In practical deployments aiming for generalization, client-specific design should be used cautiously, with an emphasis on discovering inherent global consensus~\cite{guo2023conclusion_8_1,mora2024conclusion_8_2}.}

\begin{table}
  \centering
  \caption{Performance (\%) on DP privacy preserve.}
  \label{tab: dp privacy preserve}
\begin{tabular}{c|cc|cc}
\midrule[0.3pt]
Graph-FL    & \multicolumn{2}{c|}{AIDS} & \multicolumn{2}{c}{NCI1}    \\
Simulation  & Label       & Topology   & Label         & Topology    \\ \midrule[0.3pt]
FedAvg      & 94.2±0.7    & 93.6±1.1   & 82.7±0.6      & 79.5±0.4    \\
$\epsilon$(5)-DP    & 91.9±0.9    & 90.5±1.6   & 78.3±0.7      & 74.2±0.5    \\
$\epsilon$(20)-DP   & 93.5±0.6    & 92.8±0.9   & 80.4±0.8      & 76.8±0.5    \\ \midrule[0.3pt]
GCFL        & 95.8±0.5    & 94.2±0.7   & 84.9±0.5      & 81.5±0.3    \\
$\epsilon$(5)-DP      & 92.3±0.7    & 90.8±0.9   & 81.1±0.8      & 75.1±0.4    \\
$\epsilon$(20)-DP     & 94.0±0.8    & 93.0±0.8   & 82.5±0.6      & 77.4±0.6    \\ \midrule[0.3pt]
Subgraph-FL & \multicolumn{2}{c|}{CS}   & \multicolumn{2}{c}{Physics} \\
Simulation  & Louvain+    & Metis+     & Louvain+      & Metis+      \\ \midrule[0.3pt]
FedAvg      & 83.2±0.9    & 84.0±0.7   & 91.6±0.3      & 91.0±0.4    \\
$\epsilon$(10)-DP   & 78.2±1.1    & 77.9±1.2   & 88.7±0.5      & 89.1±0.6    \\ \midrule[0.3pt]
FedGTA      & 85.0±0.7    & 85.3±0.5   & 92.8±0.4      & 91.9±0.3    \\
$\epsilon$(10)-DP   & 78.8±0.8    & 79.2±0.9   & 89.4±0.7      & 89.2±0.8    \\ \midrule[0.3pt]
FedSage+     & 84.2±1.1    & 85.6±0.7   & 92.1±0.9      & 91.4±0.5    \\
FedDEP      & 84.5±0.9    & 84.9±0.6   & 91.7±1.1      & 91.7±0.4    \\ \midrule[0.3pt]
\end{tabular}
\end{table}

\noindent
\textbf{DP-based Privacy Preserve}.
    To answer \textbf{Q6}, the experimental results are shown in Table~\ref{tab: dp privacy preserve}, where $\epsilon$ is the privacy budget. 
    To implement DP in FGL, we introduce well-selected random noise in the local model gradients for server-side perturbed model aggregation.
    More technology details can be found in~\cite{OpenFGL} (\ref{appendix: DP-based Privacy Persevere}).
    Meanwhile, we integrate FedDEP into OpenFGL, which introduces additional edge-level DP.
    Based on the results, we observe that a large $\epsilon$ enables privacy-preserving methods to match the accuracy of the non-private approach. 
    However, reducing $\epsilon$ results in a notable performance drop, primarily due to the need for local models for excessive noise injection into gradients, leading to significant degradation.
    Regarding FedDEP, compared to FedSage+, it achieves edge-level differential privacy through random sampling and enhances model capacity with a deep neighbor generation module, striking a balance between performance and privacy preservation.
    Based on this, we conclude that \textit{C9: FGL algorithms currently face a dilemma between predictive performance and privacy preservation~\cite{fu2024conclusion_9_1,zhang2024conclusion_9_2}.}

\subsection{FGL Algorithm Complexity Analysis}
\label{appendix: FGL Algorithm Complexity Analysis}
\begin{table*}[htbp]
\caption{Algorithm complexity analysis for existing prevalent FL and FGL studies. 
}
\label{tab: algorithm_analysis}
\resizebox{\textwidth}{30mm}{
\setlength{\tabcolsep}{2mm}{

\begin{tabular}{c|ccc|ccc}
\midrule[0.3pt]
Method       & Client Mem.                  & Server Mem.               & Inference Mem.        & Client Time.                  & Server Time.                      & Inference Time           \\ \midrule[0.3pt]
FedAvg       & $O((b+k)f+f^2)$              & $O(Nf^2)$                 & $O((b+k)f+f^2)$       & $O(kmf+nf^2)$                 & $O(Nf)$                           & $O(kmf+nf^2)$            \\
FedProx      & $O((b+k)f+\omega f^2)$       & $O(Nf^2)$                 & $O((b+k)f+f^2)$       & $O(kmf+nf^2+f^2)$             & $O(Nf)$                           & $O(kmf+nf^2)$            \\
Scaffold     & $O((b+k)f+\omega f^2)$       & $O(Nf^2)$                 & $O((b+k)f+f^2)$       & $O(kmf+nf^2+f^2)$             & $O(Nf+f^2)$                       & $O(kmf+nf^2)$            \\
MOON         & $O((b+k)f+Qf^2)$             & $O(Nf^2)$                 & $O((b+k)f+f^2)$       & $O(kmf+nf^2+Qnf)$             & $O(Nf)$                           & $O(kmf+nf^2)$            \\ 
FedProto     & $O((b+k)f+ cp)$              & $O(Ncp)$                  & $O((b+k)f+f^2)$       & $O(kmf+nf^2+cp^2)$            & $O(Ncp)$                          & $O(kmf+nf^2)$            \\
FedNH        & $O((b+k)f+ cp)$              & $O(N(f^2+cp)+c^2p^2)$     & $O((b+k)f+f^2)$       & $O(kmf+nf^2+cp^2)$            & $O(N(f+cp)+c^3p^3)$               & $O(kmf+nf^2)$            \\
FedTGP       & $O((b+k)f+ cp)$              & $O((N+Q)cp+p^2)$          & $O((b+k)f+f^2)$       & $O(kmf+nf^2+cp^2)$            & $O(Ncp+Qcp^2)$                    & $O(kmf+nf^2)$            \\\midrule[0.3pt]
GCFL+        & $O((b+k)f+f^2)$              & $O(TNf^2)$                & $O((b+k)f+f^2)$       & $O(kmf+nf^2)$                 & $O(N^2(\log(N)+T^2f^2))$          & $O(kmf+nf^2)$            \\
FedStar      & $O(E((b+k)f+f^2))$           & $O(Nf^2)$                 & $O(E((b+k)f+f^2))$    & $O(E(kmf+nf^2))$              & $O(Nf)$                           & $O(E(kmf+nf^2))$          \\
FedGL        & $O((b+k)f+f^2+c)$            & $O(N(f^2+nc+n^2))$        & $O((b+k)f+f^2)$       & $O(kmf+nf^2+nc^2)$            & $O(Nf+nc+n^2f^2)$                 & $O(kmf+nf^2)$ \\
FedSage+     & $O(L((n+sg)f+f^2))$          & $O(LtNf^2)$               & $O(L(n+sg)f+Lf^2)$    & $O(L((m+sg)f+(n+sg)f^2))$     & $O(Nf)$                           & $O(L((m+sg)f+(n+sg)f^2))$ \\
FGSSL        & $O(Q((b+k)f+f^2))$           & $O(Nf^2)$                 & $O(L(b+k)f+Lf^2)$     & $O(Qkmf+Qnf^2)$               & $O(Nf)$                           & $O(kmf+nf^2)$            \\
Fed-PUB      & $O(M((b+k)f+f^2)+M^2)$       & $O(N(f^2 +M) + P_g)$      & $O(M(b+k)f+Mf^2)$     & $O(Mkmf+Mnf^2)$               & $O(N^2(\log(N)+M^2))$             & $O(Mkmf+Mnf^2)$          \\
FGGP         & $O((n+sg)f+f^2+Qcp)$         & $O(Ncp)$                  & $O((b+k)f+f^2)$       & $O((m+sg)f+(n+sg)f^2+Qcp^2)$  & $O(N^2(\log(N)+c^2p^2)+Ncp)$      & $O(kmf+nf^2)$          \\
FedGTA       & $O((b+k)f+f^2+kKc)$          & $O(Nf^2+NkKc)$            & $O((b+k)f+f^2)$       & $O(km(f+knc)+n(f^2+c))$       & $O(Nf+NkKc)$                      & $O(kmf+nf^2)$            \\
AdaFGL       & $O(E((b+k)f+f^2))$           & $O(Nf^2)$                 & $O(E((b+k)f+f^2))$    & $O(E(kmf+nf^2))$              & $O(Nf)$                           & $O(E(kmf+nf^2))$           \\ 
FedTAD       & $O((b+k)f+f^2+nf)$           & $O(Nf^2+sgf+Lf^2)$         & $O((b+k)f+f^2)$       & $O(knmf+nf^2)$                & $O(Nf+(L+n)f^2+Lsgf)$             & $O(kmf+nf^2)$ \\ \midrule[0.3pt]
\end{tabular}
}}
\end{table*}
\begin{figure*}[t]
	\centering
  \includegraphics[width=\textwidth]{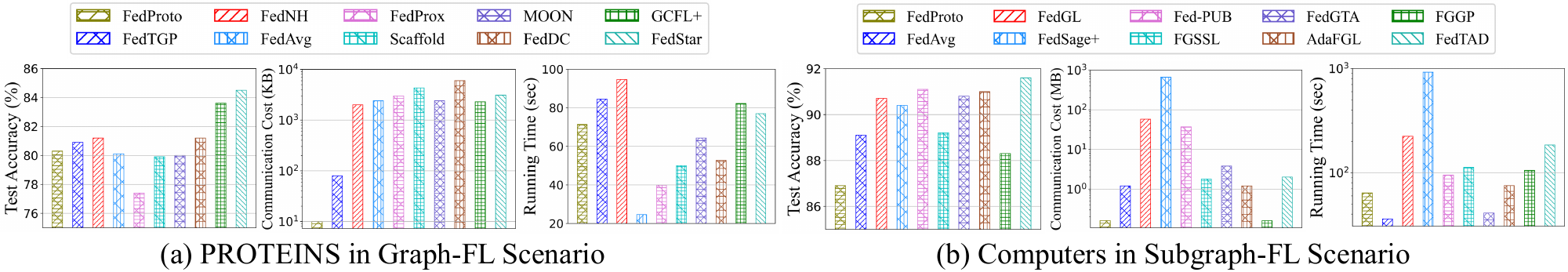}
  \caption{
    Practical efficiency in terms of performance, communication costs, and running time.
}
  \label{fig: practical efficiency}
\end{figure*}
\begin{table}
  \centering
  \caption{Performance (\%) on large-scale GraphFL.}
  \label{tab: new graph-fl dataset}
\begin{tabular}{c|cccc}
\midrule[0.3pt]
\begin{tabular}[c]{@{}c@{}}Graph-FL\\ Label (0.5)\end{tabular} & \begin{tabular}[c]{@{}c@{}}molhiv\\ (ROC-AUC)\end{tabular} & \begin{tabular}[c]{@{}c@{}}molpcba\\ (AP)\end{tabular} & \begin{tabular}[c]{@{}c@{}}ppa\\ (Acc)\end{tabular} & \begin{tabular}[c]{@{}c@{}}code2\\ (F1 score)\end{tabular} \\ \midrule[0.3pt]
FedAvg                                                         & 72.2±2.4                                                        & 20.4±0.4                                                    & 65.8±1.6                                                 & 29.8±0.4                                                        \\
FedProx                                                        & 72.8±2.2                                                        & 20.9±0.3                                                    & 65.4±1.9                                                 & 29.5±0.5                                                        \\
Scaffold                                                       & 71.7±2.5                                                        & 21.2±0.3                                                    & 66.4±1.5                                                 & 29.7±0.5                                                        \\ 
FedProto                                                       & 70.5±1.8                                                        & 19.3±0.2                                                    & 63.7±1.2                                                 & 28.6±0.3                                                        \\
FedNH                                                          & 71.2±2.0                                                        & 19.5±0.2                                                    & 64.4±1.1                                                 & 28.4±0.2                                                        \\ 
FedTGP                                                          & 70.8±2.4                                                        & 19.7±0.3                                                    & 64.2±1.4                                                 & 28.8±0.3                                                        \\\midrule[0.3pt]
GCFL+                                                          & 73.4±1.9                                                        & 21.8±0.3                                                    & 67.0±1.5                                                 & 30.5±0.3                                                        \\
FedStar                                                        & 73.8±1.7                                                        & 21.5±0.2                                                    & 67.2±1.8                                                 & 31.2±0.4                                                        \\ \midrule[0.3pt]
\end{tabular}
\end{table}
\begin{table}[htbp]  % 表格右浮动，占据半栏宽度
  \centering
  \caption{Efficiency on Louvain-based Physics (Subgraph-FL).}
  \label{tab: subgraph-fl efficiency report}
\begin{tabular}{ccccc}
\midrule[0.3pt]
Method   & Test Acc (\%) & Memory & Com. & Running Time     \\\midrule[0.3pt]
Scaffold & 90.6±0.5 & 538k   & 1076k & 152.42s \\
FedTGP   & 87.4±0.3 & 539k   & 0.38k & 76.87s  \\ \midrule[0.3pt]
FedSage+ & 91.7±0.8 & 1296k  & 1784k & 1517.96s\\
GCFL+    & 91.0±0.3 & 538k   & 538k & 184.24s  \\
Fed-PUB  & 91.6±0.5 & 1076k  & 1076k & 391.60s  \\
FedGTA   & 91.5±0.3 & 538k  & 539k  & 120.45s   \\
FedStar  & 91.3±0.5 & 1076k  & 539k & 268.12s  \\
FGSSL    & 91.1±0.6 & 751k   & 540k & 476.58s  \\
AdaFGL   & 92.0±0.4 & 964k   & 538k & 162.19s  \\ \midrule[0.3pt]
\end{tabular}
\end{table}
    To answer \textbf{Q7}, we provide a theoretical algorithm complexity analysis of the prevalent FL and FGL baselines, as illustrated in Table~\ref{tab: algorithm_analysis}, where $n$, $m$, $c$, and $f$ are the number of nodes, edges, classes, and feature dimensions, respectively. 
    $s$ is the number of selected augmented nodes and $g$ is the number of generated neighbors.
    $b$ and $T$ are the batch size and training rounds, respectively.
    $k$ and $K$ correspond to the number of times we aggregate features and moments order, respectively. 
    $N$ is the number of participating clients in each training round.
    $t$ represents the number of clients exchanging information with the current client.
    $\omega$ represents the model-wise weight alignment loss term, $Q$ denotes the size of the query set used for CL, $E$ stands for the number of models for ensemble learning, $M$ and $p$ indicate the dimension of the trainable matrix used to mask trainable weights and the prototypes.
    Besides, $P_g$ represents pseudo-graph data stored on the server side.
    
    For convenience, we choose SGC~\cite{wu2019sgc} as the local model ($k$-step feature propagation), otherwise, we adopt the model architecture ($L$-layer) used in their original paper.
    For the $k$-layer SGC model with batch size $b$, the $\mathbf{X}^{(k)}$ is the propagated feature bounded by a space complexity of $O((b+k)f)$. 
    The overhead for linear regression by multiplying $\mathbf{W}$ is $O(f^2)$.
    In the training stage, the above procedure is repeated to iteratively update the model weights. 
    For the server performing FedAvg, it needs to receive the model weights and the number of samples participating in this round.
    Its space complexity and time complexity are bounded as $O(Nf^2)$ and $O(N)$.
    As discovered by previous studies~\cite{chen2020gbp,li2024lightdic,gamlp}, the dominating term is $O(kmf)$ or $O(Lmf)$ when the graph is large since feature learning can be accelerated by parallel computation. 
    Consequently, $O(Lmf)$ emerges as the dominating complexity term of linear transformation.
    Although FGL offers a new perspective for large-scale graph learning through a distributed paradigm, it still requires the deployment of suitable scalable GNNs on the local client.

    The current mainstream trend in FL or FGL studies emphasizes the development of well-designed client-side updates to fit local data. 
    For instance, FedProx introduces model weight alignment loss, resulting in complexities of $O(\omega f^2)$. 
    Similarly, approaches like MOON, FGSSL, FGGP, FedStar, and AdaFGL employ CL loss and ensemble learning for local updates, introducing additional computational overhead upon the graph learning.
    Specifically, for the contrastive learning in MOON, FGSSL, and FGGP, the additional computational cost depends on the size and semantics of the query sample set, resulting in complexities of $O(Qf^2)$, $Q((b+k)f+f^2)$, $O(Qcp^2)$ respectively.
    This will lead to unacceptable computational overhead as the scale of local data increases. 
    As for ensemble learning approaches like FedStar and AdaFGL, which maintain multiple models locally to extract private data semantics from various perspectives, they can be bounded by $O(E((b+k)f+f^2))$.
    Furthermore, FedSage+, Fed-PUB, and FedGTA exchange additional information during communication to improve federated training.
    Despite their inherent similarities, these methods exhibit significantly different time-space complexities due to variations in their design.
    Specifically, FedSage+ involves client-side data sharing for local subgraph data augmentation, leading to a complexity of $O(L((n+sg)f+f^2))$. 
    Fed-PUB maintains a global pseudo-graph on the server side and utilizes locally uploaded weights to update trainable mask matrices for personalized learning, introducing a complexity of $O(N(f^2+M)+P_g)$. 
    In contrast, FedGTA is a lightweight method that utilizes topology-aware soft labels to encode local data, enabling personalized model aggregation on the server.
    As a result, this approach possesses a complexity of $O(kKC)$.

    While client-side training has proven effective, an increasing number of methods have recently recognized the significant potential of optimizing server-side model aggregation for federated training.
    For example, FedGL empowers local training by executing global pseudo-labeling and topological mining on the server side, which can be bounded by $O(Nf+nc+n^2f^2)$. 
    FedTAD, on the other hand, meticulously adjusts global aggregation models to enhance the initialization of local models for the next communication round through graph-specific data-free knowledge distillation, albeit incurring additional overhead of $O(Nf+(L+n)f^2+Lsgf)$. 
    
    Moreover, FedProto, FedNH, and FedTGP propose prototype-based FL. 
    They exchange class-specific embeddings between participating clients and servers in each communication round, reducing the complexity from $O(Nf^2)$ to $O(Ncp)$. 
    Additionally, FedNH optimizes global prototype initialization using interior point methods, which, while effective, poses an out-of-time risk of $O(c^3p^3)$ when datasets comprise multiple categories. 
    In comparison, FedTGP introduces independent neural architectures on the server side to adjust global prototypes, providing relative flexibility.

\begin{table*}[t]
\caption{A summary and selection suggestions for current prevalent FGL studies.}
\small
\label{tab: fgl algorithms suggestions}
\begin{tabular}{c|c|ccccc|ccc}
\midrule[0.3pt]
\multirow{2}{*}{Methods} & Effectiveness           & \multicolumn{5}{c|}{Robustness}                          & \multicolumn{3}{c}{Efficiency}            \\
                         & Statistic Heterogeneity & Noise & Sparsity & Low Client & Generalization & Privacy & Communication & Scalability & Parallelism \\ \midrule[0.3pt]
GCFL+~\cite{xie2021gcfl}                    & \textcolor{red}{\usym{2717}}                     & \textcolor{teal}{\usym{2713}}   & \textcolor{red}{\usym{2717}}      & \textcolor{teal}{\usym{2713}}        & \textcolor{red}{\usym{2717}}            & \textcolor{red}{\usym{2717}}     & \textcolor{teal}{\usym{2713}}           & \textcolor{teal}{\usym{2713}}         & \textcolor{teal}{\usym{2713}}         \\
FedStar~\cite{tan2023fedstar}                   & \textcolor{teal}{\usym{2713}}                     & \textcolor{red}{\usym{2717}}   & \textcolor{teal}{\usym{2713}}      & \textcolor{red}{\usym{2717}}        & \textcolor{teal}{\usym{2713}}            & \textcolor{red}{\usym{2717}}     & \textcolor{teal}{\usym{2713}}           & \textcolor{red}{\usym{2717}}         & \textcolor{red}{\usym{2717}}         \\
FedSage+~\cite{zhang2021fedsage}                 & \textcolor{red}{\usym{2717}}                     & \textcolor{red}{\usym{2717}}   & \textcolor{teal}{\usym{2713}}      & \textcolor{red}{\usym{2717}}        & \textcolor{red}{\usym{2717}}            & \textcolor{red}{\usym{2717}}     & \textcolor{red}{\usym{2717}}           & \textcolor{red}{\usym{2717}}         & \textcolor{red}{\usym{2717}}         \\
Fed-PUB~\cite{baek2022fedpub}                  & \textcolor{teal}{\usym{2713}}                     & \textcolor{teal}{\usym{2713}}   & \textcolor{red}{\usym{2717}}      & \textcolor{teal}{\usym{2713}}        & \textcolor{red}{\usym{2717}}            & \textcolor{red}{\usym{2717}}     & \textcolor{red}{\usym{2717}}           & \textcolor{red}{\usym{2717}}         & \textcolor{red}{\usym{2717}}         \\
FedGTA~\cite{li2023fedgta}                   & \textcolor{teal}{\usym{2713}}                     & \textcolor{teal}{\usym{2713}}   & \textcolor{red}{\usym{2717}}      & \textcolor{teal}{\usym{2713}}        & \textcolor{red}{\usym{2717}}            & \textcolor{red}{\usym{2717}}     & \textcolor{teal}{\usym{2713}}           & \textcolor{red}{\usym{2717}}         & \textcolor{teal}{\usym{2713}}         \\
FGSSL~\cite{huang2023fgssl}                    & \textcolor{red}{\usym{2717}}                     & \textcolor{red}{\usym{2717}}   & \textcolor{teal}{\usym{2713}}      & \textcolor{teal}{\usym{2713}}        & \textcolor{red}{\usym{2717}}            & \textcolor{red}{\usym{2717}}     & \textcolor{teal}{\usym{2713}}           & \textcolor{red}{\usym{2717}}         & \textcolor{red}{\usym{2717}}         \\
FedGL~\cite{chen2024fedgl}                    & \textcolor{red}{\usym{2717}}                     & \textcolor{red}{\usym{2717}}   & \textcolor{teal}{\usym{2713}}      & \textcolor{red}{\usym{2717}}        & \textcolor{teal}{\usym{2713}}            & \textcolor{red}{\usym{2717}}     & \textcolor{red}{\usym{2717}}           & \textcolor{red}{\usym{2717}}         & \textcolor{red}{\usym{2717}}         \\
AdaFGL~\cite{li2024adafgl}                   & \textcolor{teal}{\usym{2713}}                     & \textcolor{teal}{\usym{2713}}   & \textcolor{teal}{\usym{2713}}      & \textcolor{red}{\usym{2717}}        & \textcolor{teal}{\usym{2713}}            & \textcolor{red}{\usym{2717}}     & \textcolor{teal}{\usym{2713}}           & \textcolor{red}{\usym{2717}}         & \textcolor{teal}{\usym{2713}}         \\
FGGP~\cite{wan2024fggp}                     & \textcolor{teal}{\usym{2713}}                     & \textcolor{red}{\usym{2717}}   & \textcolor{red}{\usym{2717}}      & \textcolor{teal}{\usym{2713}}        & \textcolor{red}{\usym{2717}}            & \textcolor{red}{\usym{2717}}     & \textcolor{teal}{\usym{2713}}           & \textcolor{teal}{\usym{2713}}         & \textcolor{red}{\usym{2717}}         \\
FedDEP~\cite{zhang2024feddep}                   & \textcolor{red}{\usym{2717}}                     & \textcolor{red}{\usym{2717}}   & \textcolor{teal}{\usym{2713}}      & \textcolor{red}{\usym{2717}}        & \textcolor{red}{\usym{2717}}            & \textcolor{teal}{\usym{2713}}     & \textcolor{teal}{\usym{2713}}           & \textcolor{red}{\usym{2717}}         & \textcolor{red}{\usym{2717}}         \\
FedTAD~\cite{zhu2024fedtad}                   & \textcolor{teal}{\usym{2713}}                     & \textcolor{red}{\usym{2717}}   & \textcolor{teal}{\usym{2713}}      & \textcolor{red}{\usym{2717}}        & \textcolor{teal}{\usym{2713}}            & \textcolor{red}{\usym{2717}}     & \textcolor{teal}{\usym{2713}}           & \textcolor{red}{\usym{2717}}         & \textcolor{red}{\usym{2717}}         \\ \midrule[0.3pt]
\end{tabular}
\end{table*}
    
\subsection{Efficiency Evaluation}
\label{sec: Efficiency Evaluation}
    To answer \textbf{Q8}, we provide the efficiency reports in Fig.~\ref{fig: practical efficiency}, Table~\ref{tab: new graph-fl dataset}, and Table~\ref{tab: subgraph-fl efficiency report}, and obtain the following observations:
    (1) Prototype-based FedProto, FedTGP, and FGGP reduce communication costs but require extra computation, offsetting their runtime advantage.
    (2) Cross-client FedGL and FedSage+ suffer reduced efficiency due to delays from inter-client communication.
    (3) Decoupled AdaFGL maximizes local computation and minimizes communication costs, providing efficiency advantages.
    (4) In Graph-FL, models treat data samples independently, so even with large-scale graph samples, the limited size of each graph minimizes OOM or OOT issues.  
    However, no single approach consistently achieves superior performance.
    (5) In our experiments, Param. refers to trainable parameters, Time to total runtime, and Com. to communication costs. 
    FedSage+ and AdaFGL perform better but incur high Com. and computational complexity.
    Consequently, we conclude that \textit{C10: FGL algorithms leveraging prototypes and decoupled techniques (i.e., multi-client collaboration then local updates) demonstrate substantial potential in applications with stringent efficiency requirements~\cite{heinbaugh2023conclusion_10_1,li2024conclusion_10_2}.}

\subsection{FGL Guidance and OpenFGL Tutorial}
\label{sec: OpenFGL Algorithm Tutorial}
In this section, we provide a detailed overview of OpenFGL:
(1) Comprehensive evaluation of FGL methods to guide deployment (FGL Guidance);
(2) User-friendly APIs to facilitate the reproduction and further development of FGL algorithms (OpenFGL Tutorial).

\noindent
\textbf{FGL Guidance.}
    In Sec.\ref{sec: Performance Comparison}-Sec.\ref{sec: Efficiency Evaluation}, we conduct a comprehensive empirical investigation of effectiveness, robustness, and efficiency, drawing 10 conclusions. 
    These insights are crucial for selecting appropriate FGL algorithms in real-world applications.
    To provide a clearer presentation, we summarize the prevalent FGL algorithms in Table~\ref{tab: fgl algorithms suggestions}. 
    Notably, scalability refers to the ability of a method to handle large-scale graphs without OOM or OOT. 
    Regarding parallelism, we evaluate whether the method relies heavily on server-broadcasted information for local training. 
    Minimal dependency allows for fewer communication and enables more independent parallel client training.
    Based on this, we observe that current FGL algorithms struggle to maintain consistent competitiveness across various requirements, highlighting that the field is still in its early stages with significant potential for future development.

\noindent
\textbf{OpenFGL Tutorial.}
    We now present the algorithm design principles, which offer a unified API. 
    It uses the {\textit{FGLTrainer}} to manage client-server communication during each training round, aggregating messages via the \textit{message\_pool} variable. 
    Users can customize the {\textit{FGLClient}} and {\textit{FGLServer}} to adjust message content and rules for specific algorithms. 
    For clarity, we provide \textit{PyTorch-style} implementations, exemplified by the \textit{FedAvg} algorithm:

\begin{algorithm}[t]
\caption{\textbf{OpenFGL}-\textbf{\textit{FGLTrainer}}.\textit{Pytorch style}.}
\label{alg:trainer}
\definecolor{codeblue}{rgb}{0.25,0.5,0.5}
\lstset{
  backgroundcolor=\color{white},
  basicstyle=\fontsize{7.2pt}{7.2pt}\ttfamily\selectfont,
  columns=fullflexible,
  breaklines=true,
  captionpos=b,
  commentstyle=\fontsize{7.2pt}{7.2pt}\color{codeblue},
  keywordstyle=\fontsize{7.2pt}{7.2pt},
% ##  frame=tb,
}
\begin{lstlisting}[language=python]
class FGLTrainer:
    def __init__(self, args):
        self.args = args
        self.message_pool = {}
        self.clients = load_client(args,...)
      
    def train(self):
        for round_id in range(self.args.num_rounds):
            self.message_pool["round"] = round_id
            self.message_pool["sampled_clients"] = sampled_clients
            self.server.send_message()
            for client_id in sampled_clients:
                self.clients[client_id].execute()
                self.clients[client_id].send_message()
            self.server.execute()
            self.evaluate()  
\end{lstlisting}
\end{algorithm}
\noindent
    \textit{(a) FGLTrainer.} 
    This class manages message and command flows between clients and a central server. 
    In each training round, the trainer selects clients, updates the \textit{message\_pool}, and dispatches server messages.
    Clients process tasks locally and send updates to the server for global aggregation. 
    Each round includes an evaluation phase. 
    The implementation of {\textit{FGLTrainer}} is shown in Algorithm~\ref{alg:trainer}.

\noindent
\begin{algorithm}[t]
\caption{\textbf{OpenFGL}-\textbf{\textit{FGLClient}}.\textit{Pytorch style}.}
\label{alg:client}
\definecolor{codeblue}{rgb}{0.25,0.5,0.5}
\lstset{
  backgroundcolor=\color{white},
  basicstyle=\fontsize{7.2pt}{7.2pt}\ttfamily\selectfont,
  columns=fullflexible,
  breaklines=true,
  captionpos=b,
  commentstyle=\fontsize{7.2pt}{7.2pt}\color{codeblue},
  keywordstyle=\fontsize{7.2pt}{7.2pt},
% ##  frame=tb,
}
\begin{lstlisting}[language=python]
class FedAvgClient(BaseClient):
    def __init__(self, args, client_id...):   
        
    def execute(self):
        # Receive messages (global model weights) from the server stored in the message pool to update the local model
        with torch.no_grad():
            for (l_param, g_param) in zip(self.task.model.parameters(), self.message_pool[server][weight]):
                local_param.data.copy_(global_param)
        self.task.train()

    def send_message(self):
        self.message_pool[f"client_{self.client_id}"] = {
                num_samples: self.task.num_samples,
                weight: list(self.task.model.parameters())}
\end{lstlisting}
\end{algorithm}
    \textit{(b) FedAvgClient.} 
    Users only need to customize:
    (1) Local execution, where FedAvg downloads the global model and performs local training;
    (2) Client-to-server messages, including the local sample count and model weights in FedAvg.
    The \textit{Pytorch-style} implementation of {\textit{FedAvgClient}} is provided in Algorithm~\ref{alg:client}.

\noindent
\begin{algorithm}[t]
\caption{\textbf{OpenFGL}-\textbf{\textit{FGLServer}}: \textit{Pytorch style}.}
\label{alg:server}
\definecolor{codeblue}{rgb}{0.25,0.5,0.5}
\lstset{
  backgroundcolor=\color{white},
  basicstyle=\fontsize{7.2pt}{7.2pt}\ttfamily\selectfont,
  columns=fullflexible,
  breaklines=true,
  captionpos=b,
  commentstyle=\fontsize{7.2pt}{7.2pt}\color{codeblue},
  keywordstyle=\fontsize{7.2pt}{7.2pt},
% ##  frame=tb,
}
\begin{lstlisting}[language=python]
class FedAvgServer(BaseServer):
    def __init__(self, args, message_pool...):
   
    def execute(self):
        # Receive messages (number of samples, local model weight) from each client stored in the message pool
        with torch.no_grad():
            for it, client_id in enumerate(sampled_clients):
                weight = self.message_pool[client_id][num_samples]/num_tot_samples (FedAvg_model_aggregation...)
        
    def send_message(self):
        self.message_pool["server"] = {
            weight: list(self.task.model.parameters())}
\end{lstlisting}
\end{algorithm}
    \textit{(c) FedAvgServer.}
    Similar to {\textit{FedAvgClient}}, users customize two modules:
    (1) Global execution (e.g., in FedAvg: weighted model averaging based on sample size);
    (2) Server-to-client messages (e.g., in FedAvg: global model weights).
    The \textit{Pytorch-style} implementation of {\textit{FedAvgServer}} is illustrated in Algorithm~\ref{alg:server}.

\section{Conclusion and Future Directions}
\label{sec: Conclusion and Future Directions}
    In this paper, we first present a comprehensive overview of the current research progress in the FGL field and the significant potential of this technology for deployment in graph-based database applications.
    Subsequently, we propose OpenFGL, a comprehensive FGL benchmark, which encompasses 18 recently proposed SOTA FGL algorithms and 42 datasets from 18 domains for 5 downstream tasks across 2 prevalent FGL scenarios. 
    The goal of our work is to fairly examine the current state of FGL development and offer key insights for future research endeavors.
    Although FGL primarily serves downstream tasks in graph-based ML, it also holds significant potential for database applications.
    Specifically, it enables each client to generate high-quality graph embeddings (nodes, edges, and subgraphs) in a privacy-preserve and distributed manner.
    These embeddings provide rich semantic representations, which can be leveraged for efficient retrieval in vector databases. 
    This perspective introduces new opportunities for applying FGL techniques to enhance the performance and scalability of graph-based databases.

    Subsequently, we conduct extensive experiments aimed at unveiling the performance of FGL algorithms from 3 orthogonal perspectives: effectiveness, robustness, and efficiency.
    Our investigations reveal promising advancements achieved by FGL studies but also highlight their limitations, such as vulnerability in inadequate node descriptions, robustness, and scalability.
    To inspire future research, we combine experimental \textit{\textbf{Conclusions}} to present the following significant challenges and promising directions.

    For \textbf{Effectiveness}, 
    (1) \textbf{Quantify Distributed Graphs} (\textit{C1} and \textit{C3}).
    Essentially, the potential benefits of FGL in real-world deployments stem from the uniform graph distribution. 
    However, the entanglement of node features, labels, and topology poses a challenge in explicitly quantifying statistics within distributed graphs, resulting in coarse descriptions. 
    This constraint sharply contrasts with the intuitive semantic feature and label distribution skew observed in CV-based FL. 
    Therefore, quantifying the statistics of multi-source graphs is crucial.
    (2) \textbf{FGL Heterogeneity} (\textit{C2} and \textit{C4}).
    Although some FGL studies attempt to define graph-based heterogeneity challenges, these definitions are often insufficient due to the complexity of graph characteristics and the diversity of applications.
    Furthermore, the advantages of these FGL algorithms (e.g., AdaFGL) lack significant impact. 
    Consequently, there is still necessary effort to be made in addressing FGL heterogeneity.

    For \textbf{Robustness}, 
    (3) \textbf{Personalized FGL} (\textit{C5} and \textit{C7}).
    In real-world scenarios, the robustness of FGL against client-specific noise and low client participation in communication is essential. 
    During federated training, these factors significantly impact the attainment of global consensus, thereby hindering high-quality supervision provided for local training.
    Fortunately, personalized techniques can leverage local reliable knowledge to establish unbiased global consensus and customize broadcasts for each client, guiding local updates.
    This mitigates optimization challenges arising from noise data and limited data sources. 
    Consequently, personalized FGL algorithms emerge as an effective strategy for addressing these concerns.
    (4) \textbf{Multi-client Collaboration FGL} (\textit{C6} and \textit{C8}).
    During our investigation, we found that promoting server-side multi-client collaboration can extract global insights from sparse data.
    Additionally, this collaborative approach can capture shared semantic knowledge across data domains to facilitate robust generalization. 
    Therefore, there is an emergency demand for more efficient multi-client collaboration FGL algorithms.
    (5) \textbf{Privacy-preserve FGL} (\textit{C9}).
    The goal of FGL is to achieve multi-client collaborative training in a privacy-preserving manner without direct data sharing.
    However, current FGL algorithms, in pursuit of superior performance, increasingly share local information, raising potential concerns. 
    Therefore, the development of FGL algorithms with strict privacy requirements is imperative.
    Meanwhile, more privacy-preserving technologies should be considered for further explorations.

    For \textbf{Efficiency},
    (6) \textbf{Decoupled and Scalable FGL} (\textit{C10}).
    Existing FGL algorithms face challenges in practical deployment due to communication delay and topology mining, making it difficult to handle large-scale datasets. 
    Therefore, developing new federated collaboration paradigms such as decoupled mechanisms and focusing on algorithm design scalability is crucial.

    FGL should establish federated collaboration standards for various graph types (e.g., directed, signed, hypergraphs, heterogeneous) and learning paradigms (e.g., unsupervised, few-shot, continual, unlearning) based on the data systems~\cite{tian2023gnn_database2,zhou2023gnn_database3,besta2022gnn_database1}. 
    However, FGL remains a burgeoning field with numerous research gaps. 
    Nevertheless, we are committed to continually enhancing OpenFGL to support future research endeavors. 
    For instance, we are progressively refining the execution standards for federated heterogeneous graph learning. 
    Notably, considering the space constraints and the need for a clear and reader-friendly presentation, we provide an overview and corresponding experimental results in ~\cite{OpenFGL} (\ref{appendix: Federated Heterogeneous Graph Learning}).

\clearpage

\balance{
\bibliographystyle{ACM-Reference-Format}
\bibliography{reference}
}

\clearpage
\appendix

\section{Outline}

The appendix is organized as follows:
\begin{description}
    \item[A.1] Dataset Description.
    % \item[A.2] FGL Scenario Simulation Description.
    \item[A.2] Baseline Description.
    \item[A.3] Metric Description.
    \item[A.4] Robustness Simulation Description.
    \item[A.5] Effectiveness Evaluation Strategies.
    \item[A.6] Experiment Environment.
    \item[A.7] Hyperparameter Settings.
    \item[A.8] DP-based Privacy Persevere.
    \item[A.9] Federated Heterogeneous Graph Learning.
\end{description}

\subsection{Dataset Description}
\label{appendix: Dataset Description}
    In Graph-FL, we conduct experiments on the compounds networks (MUTAG, BZR, COX2, DHFR, PTC-MR, AIDS, NCI1, hERG, ogbg-molhiv, ogbg-molpcba)~\cite{debnath1991_MUTAG, sutherland2003_COX2_BZR_DHFR, helma2001_PTC_PROTEINS, riesen2008_AIDS, wale2008_NCI1,gaulton2017_hERG, hu2020ogb}, protein networks (ENZYMES, DD, PROTEINS, ogbg-ppa)~\cite{borgwardt2005_ENZYMES, dobson2003_DD, helma2001_PTC_PROTEINS, hu2020ogb}, collaboration network (COLLAB)~\cite{leskovec2005graphs_COLLAB}, movie networks (IMDB-BINARY, IMDB-MULTI)~\cite{yanardag2015_IMDB_B_M}, super-pixel networks (MNISTSuperPixels)~\cite{monti2017geometric}, point cloud networks (ShapeNet)~\cite{yi2016scalable}, and yntax trees (ogbg-code2)~\cite{hu2020ogb}.
    In Graph-FL, node features generally include both node attributes and labels, which provide essential information for characterizing individual nodes within a graph. 
    In most graph-level GNNs, these components are concatenated and subsequently fed into the model. 
    As a result, we report the feature dimension as the combined size of both node attributes and labels.
    Furthermore, super-pixel networks provide an advanced approach to representing images as graphs, where nodes correspond to super-pixels, and edges capture spatial or semantic relationships between these regions.
    This graph-based representation effectively harnesses both the structural and semantic properties of the image, resulting in enhanced classification performance by moving beyond simple pixel-level analysis.
    A detailed description of the datasets used in our experiments is provided below:

    \textbf{MUTAG}~\cite{debnath1991_MUTAG} is a widely used bioinformatics dataset consisting of 188 graphs, each representing a nitro compound. 
    The nodes are labeled with one of 7 distinct node labels.
    The primary objective of this dataset is to classify each graph to determine whether the corresponding compound is mutagenic, specifically distinguishing between aromatic and heteroaromatic compounds.

    \textbf{BZR}~\cite{sutherland2003_COX2_BZR_DHFR} is a bioinformatics dataset aimed at predicting compound activity, with a primary focus on a diverse collection of benzimidazole compounds. 
    This dataset is designed to provide information on the concentration required for each compound to inhibit the activity of specific biomolecules, thereby offering valuable insights into their effectiveness in modulating biological processes. 
    Such data are critical for understanding the biochemical interactions of benzimidazole compounds and can serve as a resource for developing novel therapeutics by identifying promising candidates for further exploration in drug discovery.

    \textbf{COX2}~\cite{sutherland2003_COX2_BZR_DHFR} is a dataset focused on Cyclooxygenase-2, an enzyme that plays a crucial role in mediating inflammation and pain. 
    The dataset is used to classify a variety of chemical compounds and predict their potential inhibitory potency against the COX-2 enzyme.
    Given that COX-2 is a key target in the development of anti-inflammatory drugs, this dataset is highly valuable for computational chemistry and drug discovery research, facilitating the identification of promising candidate compounds for therapeutic intervention in inflammatory conditions.

    \textbf{DHFR}~\cite{sutherland2003_COX2_BZR_DHFR} describes the inhibitory activity of compounds against dihydrofolate reductase, an enzyme critical in cellular folate metabolism and a target for many anticancers.
    By representing compound molecules in the DHFR dataset as graphs, researchers can employ machine learning models to predict the potential inhibitory effects of unknown compounds against dihydrofolate reductase.

    \textbf{PTC-MR}~\cite{helma2001_PTC_PROTEINS} is a bioinformatics dataset consisting of 344 graphs, each representing a chemical compound. 
    Within these graphs, nodes are labeled with one of 19 distinct node labels. 
    The primary objective of this dataset is to predict the carcinogenicity of each compound in rodents, making it valuable for studies related to chemical toxicity and safety assessments.

    \textbf{AIDS}~\cite{riesen2008_AIDS} is a graph dataset comprising 2000 graphs, each representing molecular compounds derived from the AIDS Antiviral Screen Database of Active Compounds. 
    The dataset includes a total of 4395 chemical compounds, categorized into three classes: 423 compounds belonging to class CA, 1081 to class CM, and the remaining compounds to class CI. 
    This dataset is widely used for tasks involving molecular classification and drug discovery research.

    \textbf{NCI1}~\cite{wale2008_NCI1} is a bioinformatics dataset comprising 4,110 graphs representing chemical compounds. 
    It contains data published by the National Cancer Institute (NCI). 
    Each node is assigned with one of 37 discrete node labels. 
    The graph classification label is determined by NCI anti-cancer screens assessing the ability to suppress or inhibit the growth of a panel of human tumor cell lines.

    \textbf{hERG}~\cite{gaulton2017_hERG} consists of molecular graphs that represent atoms and chemical bonds within various compounds. 
    This dataset is crucial for predicting the inhibitory effects of these compounds on the human Ether-à-go-go-Related Gene (hERG) potassium channel, which is important in drug safety assessment.
    The prediction task associated with hERG is essentially a graph regression, making it a valuable resource for studies in drug discovery and toxicology.

    \textbf{ogbg-molhiv, ogbg-molpcba}~\cite{hu2020ogb} are molecular property prediction datasets of different sizes: ogbg-molhiv (small) and ogbg-molpcba (medium). 
    Both are derived from MoleculeNet and are among its largest datasets. 
    Molecules are pre-processed with RDKit, where each graph represents a molecule with nodes as atoms and edges as chemical bonds. 
    Input node features are 9-dimensional, including atomic number, chirality, and other properties. 
    Full feature descriptions are in the code. 
    A script to convert SMILES strings into graph objects, which requires RDKit, can pre-process external molecular datasets for compatibility with OGB datasets.

    \textbf{ENZYMES}~\cite{borgwardt2005_ENZYMES} is a comprehensive dataset consisting of 600 protein tertiary structures, meticulously curated from the BRENDA enzyme database. 
    This dataset provides researchers with the opportunity to study the intricate and diverse structures of six distinct enzyme classes, enabling in-depth computational analysis and facilitating the development of advanced machine learning models for biological research.
    The rich structural diversity captured in ENZYMES makes it a valuable resource for tasks such as protein classification and function prediction, and other bioinformatics applications, ultimately contributing to a deeper understanding of enzyme mechanisms.

    \textbf{DD}~\cite{dobson2003_DD} is a bioinformatics dataset composed of 1,178 graph structures representing proteins. 
    In these graphs, nodes correspond to amino acids, and edges connect nodes that are within 6 Angstroms of each other, reflecting the spatial proximity of amino acids within the protein structure. 
    The primary task associated with this dataset is a binary classification to differentiate between enzymes and non-enzymes, making it a valuable resource for studies in protein function prediction and structural bioinformatics.

    \textbf{PROTEINS}~\cite{helma2001_PTC_PROTEINS} is a bioinformatics dataset comprising 1,113 structured proteins. 
    Nodes in these graph-based proteins denote secondary structure elements and are assigned discrete node labels indicating whether they represent a helix, sheet, or turn. 
    Edges indicate adjacency along the amino-acid sequence or in space between two nodes. 
    The objective is to predict the protein function.

    \textbf{ogbg-ppa}~\cite{hu2020ogb} comprises undirected protein association neighborhoods extracted from networks of 1,581 species, covering 37 taxonomic groups (e.g., mammals, bacteria, archaeans) across the tree of life.
    To construct these neighborhoods, 100 proteins were randomly selected from each species, and 2-hop neighborhoods were built. 
    The center node was removed, and neighborhoods were sub-sampled to ensure fewer than 300 nodes. 
    Nodes represent proteins, and edges denote biologically meaningful associations, with 7-dimensional features indicating the confidence (0-1) of specific associations, such as gene co-occurrence, gene fusion, and co-expression.
    
    \textbf{COLLAB}~\cite{leskovec2005graphs_COLLAB} is a scientific collaboration dataset consisting of 5,000 ego networks represented as graphs. 
    This dataset is compiled from three public collaboration datasets. 
    Each ego network comprises researchers from various fields and is labeled according to the corresponding field, namely High Energy Physics, Condensed Matter Physics, and Astrophysics.

    \textbf{IMDB-BINARY}~\cite{yanardag2015_IMDB_B_M} is a movie collaboration dataset comprising 1,000 graphs representing ego networks for actors and actresses.
    Derived from collaboration graphs within the Action and Romance genres, each graph features nodes representing actors/actresses and edges denoting their collaboration in the same movie. 
    Graphs are labeled according to the corresponding genre, and the objective is to classify the genre for each graph.

    \textbf{IMDB-MULTI}~\cite{yanardag2015_IMDB_B_M} is the multi-class extension of the IMDB-BINARY dataset, comprising 1,500 ego-networks. 
    It includes three additional movie genres: Comedy, Romance, and Sci-Fi, making it suitable for multi-class classification tasks. 
    This dataset is commonly used to evaluate the performance of graph-level algorithms.

    \textbf{MNISTSuperPixels}~\cite{monti2017geometric} is a super-pixel network dataset derived from the original MNIST dataset, where the standard 28x28 pixel images are converted into graphs with 75 nodes each. In this transformation, each node represents a super-pixel, which is a cluster of nearby pixels grouped based on their intensity and spatial proximity. The edges between nodes capture the spatial relationships between these super-pixels.

    \textbf{ShapeNet}~\cite{yi2016scalable} is a comprehensive point cloud dataset consisting of approximately 17,000 3D shape point clouds spanning 16 different shape categories.
    Each category is further annotated with 2 to 6 parts, providing detailed segmentation labels. 
    The primary objectives for utilizing this dataset are point cloud classification and segmentation, making it an essential resource for benchmarking algorithms in 3D shape analysis, recognition, and understanding.

    \textbf{ogbg-code2}~\cite{hu2020ogb} comprises abstract syntax trees derived from approximately 450,000 Python method definitions. 
    These methods are extracted from 13,587 repositories across popular GitHub projects. 
    The dataset originates from GitHub CodeSearchNet, a collection of datasets and benchmarks for machine learning-based code retrieval. 
    In ogbg-code2, we add an additional feature extraction step, including AST edges, AST nodes, and tokenized method names. 
    This enables ogbg-code2 to capture the source code's graph structure, beyond its token sequence representation.
    
    As for the Subgraph-FL, OpenFGL integrates citation networks (Cora, Citeseer, PubMed, FedDBLP, ogbn-arxiv)~\cite{Yang16cora, WangFedScope_22_fsg, hu2020ogb}, co-purchase networks (Amazon-Computers, Amazon-Photo, ogbn-products)~\cite{shchur2018amazon_datasets, hu2020ogb}, co-author networks (Co-author CS, Co-author Physics)~\cite{shchur2018amazon_datasets}, wiki-page networks (Chameleon, ChameleonFilter, Squirrel, SquirrelFilter)~\cite{pei2020geomgcn, platonov2023hete_gnn_survey4}, actor network (Actor)~\cite{pei2020geomgcn}, game synthetic network (Minesweeper)~\cite{platonov2023hete_gnn_survey4}, crowd-sourcing network (Tolokers)~\cite{platonov2023hete_gnn_survey4}, article syntax network (Roman-empire)~\cite{platonov2023hete_gnn_survey4}, rating network (Amazon-rating)~\cite{platonov2023hete_gnn_survey4}, social network (Questions)~\cite{platonov2023hete_gnn_survey4}, and point cloud networks (PCPNet, S3DIS)~\cite{guerrero2018pcpnet,armeni20163d}.
    
    Remarkably, while recent research has highlighted potential data leakage issues due to duplicates in the original Chameleon and Squirrel datasets, considering that previous studies commonly utilized the original versions for validation, we integrated both versions in OpenFGL to offer comprehensive evaluation.
    Furthermore, point cloud datasets are utilized for downstream tasks where each point is associated with geometric attributes like surface normals and curvature, serving as node features in the graph representation. 
    For surface normal estimation, GNNs are employed to treat each point as a node in the k-nearest neighbors (KNN) graph, learning to predict normal vectors by aggregating information from neighboring points. 
    Similarly, in curvature estimation, GNNs capture local geometric features to predict curvature values.
    For point cloud classification, the entire point cloud is represented as a graph, with GNNs aggregating point-level information to classify the cloud. 
    In point cloud segmentation, GNNs assign labels to each node, segmenting the cloud based on local and global context. 
    These methods effectively utilize GNNs to leverage the geometric relationships within point clouds for tasks like regression, classification, and segmentation.
    Developers can flexibly use the above point cloud data.
    The detailed description of Subgraph-FL datasets is listed below:

    \textbf{Cora}, \textbf{CiteSeer}, and \textbf{PubMed}~\cite{Yang16cora} are widely used citation network datasets, where nodes represent papers and edges denote citation relationships.
    Node features are word vectors, indicating the presence or absence of specific words in each paper. 
    These datasets are frequently used for node classification.

    \textbf{FedDBLP}~\cite{WangFedScope_22_fsg} is the first collected dataset in a distributed manner, where each node represents a published paper and each edge signifies a citation.
    The bag of words from each paper’s abstract is used as node attributes, and the paper's theme is designated as its label. 
    To simulate scenarios where a venue or organizer restricts citations of its papers, users can split the dataset based on each node’s venue or the organizer of that venue.

    \textbf{ogbn-arxiv}~\cite{hu2020ogb} is a widely used citation graph indexed by Microsoft Academic Graph (MAG)~\cite{wang2020microsoft_MAG}, especially for the large-scale graph learning. 
    Each paper in the dataset is represented by the average of the word embeddings derived from its title and abstract. 
    These word embeddings are generated using the skip-gram model, which captures semantic relationships between words based on their context within the text.
    This dataset is widely used for graph-based learning tasks, such as node classification.

    \textbf{Amazon Photo} and \textbf{Amazon Computers}~\cite{shchur2018amazon_datasets} are subsets of the Amazon co-purchase graph, where nodes represent individual products, and edges signify that two products are frequently bought together. 
    The node features for these datasets are derived from product reviews, represented as bag-of-words vectors, capturing the textual information associated with each item. 
    These datasets are commonly used for graph-based downstream tasks such as node classification in graph-based recommendation systems.

    \textbf{ogbn-products}~\cite{hu2020ogb} is a co-purchasing network where nodes represent products and edges indicate frequent co-purchases. 
    The node features are derived from bag-of-words representations of product descriptions. 
    Due to its extensive size and complex structure, this dataset is particularly well-suited for large-scale graph learning applications, making it an ideal benchmark for evaluating the scalability and performance of graph-based algorithms.

    \textbf{Coauthor CS} and \textbf{Coauthor Physics}~\cite{shchur2018amazon_datasets} are co-authorship graphs derived from the MAG~\cite{wang2020microsoft_MAG}. 
    In these graphs, nodes represent individual authors, edges denote co-authorship relationships between them, and node features are constructed from the keywords of the authors' publications. 
    The labels assigned to the nodes indicate the specific research fields in which the authors are active. 
    These datasets are commonly used for evaluating graph-based methods, particularly in the context of node classification.

    \textbf{Chameleon} and \textbf{Squirrel}~\cite{pei2020geomgcn} are two page-page networks extracted from specific topics within Wikipedia. 
    In these datasets, nodes represent web pages, while edges signify mutual links between pages. 
    Node features are derived from several informative nouns found on Wikipedia. 
    They categorize the nodes into five groups based on the average monthly web page traffic.

    \textbf{Chameleon Filter} and \textbf{Squirrel Filter}~\cite{platonov2023hete_gnn_survey4} emphasis nodes in original datasets share the same regression target and neighborhood simultaneously, resulting in duplicates.
    These duplicates are present across the training, validation, and test sets, causing data leakage. 
    Therefore, these filtered versions enable a fairer comparison.

    \textbf{Actor}~\cite{pei2020geomgcn} is an actor co-occurrence network where nodes represent actors, and edges indicate their co-appearance on Wikipedia pages. 
    Node features are bag-of-words vectors derived from these pages, and actors are categorized into five groups based on the terms found in their respective Wikipedia entries. 
    This dataset is commonly used for graph-based tasks like node classification.

    \textbf{Minesweeper}~\cite{platonov2023hete_gnn_survey4} draws inspiration from the Minesweeper game and stands as the synthetic dataset. 
    The graph is a regular 100x100 grid, where each node (cell) is linked to its eight neighboring nodes (excluding nodes at the grid's edge, which have fewer neighbors). 
    Twenty percent of the nodes are randomly designated as mines. 
    The objective is to predict which nodes conceal mines. 
    Node features consist of one-hot-encoded counts of neighboring mines. 
    However, for a randomly chosen 50\% of the nodes, the features are undisclosed, indicated by a distinct binary feature.

    \textbf{Tolokers}~\cite{platonov2023hete_gnn_survey4} is derived from the crowdsourcing platform~\cite{Tolokers_original}.
    Nodes correspond to workers who have engaged in at least one of the 13 selected projects. 
    An edge connects two workers if they have collaborated on the same task. 
    The objective is to predict which workers have been banned in one of the projects.

    \textbf{Roman-empire}~\cite{platonov2023hete_gnn_survey4} is based on the Roman Empire article from the English Wikipedia~\cite{lhoest2021empire_original}, each node corresponds to a non-unique word in the text, mirroring the article's length. 
    Nodes are connected by an edge if the words either follow each other in the text or are linked in the sentence's dependency tree. 
    Thus, the graph represents a chain graph with additional connections.

    \textbf{Amazon-ratings}~\cite{platonov2023hete_gnn_survey4} is derived from the co-purchasing network and its metadata available in the SNAP~\cite{leskovec2014rating_original}. 
    Nodes are items and edges connect items frequently bought together. 
    The task is predicting the average rating given by reviewers, categorized into five classes. 
    Node features are based on the FastText embeddings~\cite{grave2018fast_word_embedding} of words in the product description.
    To manage graph size, only the largest connected component of the 5-core is considered.

    \textbf{Questions}~\cite{platonov2023hete_gnn_survey4} is derived from data collected from the question-answering platform Yandex Q. 
    In this dataset, nodes represent users, and an edge exists between two nodes if one user answers another user's question within a one-year timeframe (from September 2021 to August 2022). 
    The objective is to predict which users remained active on the website (i.e., were not deleted or blocked) by the end of the specified period. 
    For node features, it utilizes the average FastText embeddings for words found in the user descriptions. 

    \textbf{PCPNet}~\cite{guerrero2018pcpnet} is a point cloud dataset consisting of 30 distinct shapes, each represented as a densely sampled point cloud with 100,000 points. 
    For each shape, surface normals and local curvatures are provided as node features, capturing essential geometric properties.
    This dataset is intended for tasks such as point cloud classification and segmentation, making it a valuable resource for evaluating algorithms in graph learning and 3D shape analysis.
    
    \textbf{S3DIS}~\cite{armeni20163d} is a point cloud dataset comprising six large-scale indoor areas from three different buildings. 
    It includes 12 semantic elements, such as walls, floors, and furniture, as well as one clutter class, making it a diverse dataset for indoor scene understanding. 
    The primary objectives for using this dataset are point cloud classification and segmentation, offering a challenging benchmark for evaluating the performance of algorithms in recognizing and segmenting complex indoor environments.

%%%%%%%%%%%%%%%%%%%%%%%%%%%%%%%%%%%%%%%%%%%%%%%%%%%%%%%%%%%%

% \subsection{FGL Scenario Simulation Description}\label{appendix: FGL Scenario Simulation Description}

\vspace{0.22cm}

\subsection{Baseline Description}
\label{appendix: Baseline Description}
    Given the uniqueness of FGL, the baseline of OpenFGL consists of three components:
    (1) Backbone graph learning models in multi-clients;
    (2) Prevalent FL algorithm in graph-independent scenarios (i.e., computer vision);
    (3) Recently proposed FGL algorithms.
    
    For (1), considering that most FGL approaches entail additional design for the local backbone model, we have implemented only the most popular baseline models (GCN~\cite{kipf2016gcn}, GAT~\cite{velivckovic2017gat}, GraphSAGE~\cite{hamilton2017graphsage}, SGC~\cite{wu2019sgc}, GCNII~\cite{chen2020gcnii}, GIN~\cite{xu2018gin}, TopKPooling~\cite{gao2019_TopKPooling}, SAGPooling~\cite{lee2019_SAGPooling}, EdgePooling~\cite{diehl2019_EdgePooling}, and PANPooling~\cite{ma2020_PANPooling}) in centralized graph learning, which is generally applicable to both graph-FL and subgraph-FL scenarios, providing flexibility to the future FGL developers.
    The backbone GNN details implemented in our proposed OpenFGL are listed below:

    \textbf{GCN}~\cite{kipf2016gcn} introduces a novel approach to graphs that is based on a first-order approximation of spectral convolutions on graphs.
    This approach learns hidden layer representations that encode both local graph structure and features of nodes.

    \textbf{GAT}~\cite{velivckovic2017gat} utilizes attention mechanisms to quantify the importance of neighbors for message aggregation.
    This strategy enables implicitly specifying different weights to different nodes in a neighborhood, without depending on the graph structure upfront.

    \textbf{GraphSAGE}~\cite{hamilton2017graphsage} leverages neighbor node attribute information to efficiently generate representations.
    This method introduces a general inductive framework that leverages node feature information to generate node embeddings for previously unseen data. 

    \textbf{SGC}~\cite{wu2019sgc} simplifies GCN by removing non-linearities and collapsing weight matrices between consecutive layers.
    Theoretical analysis show that the simplified model corresponds to a fixed low-pass filter followed by a linear classifier.
    
    \textbf{GCNII}~\cite{chen2020gcnii} incorporates initial residual and identity mapping. Theoretical and empirical evidence is presented to demonstrate how these techniques alleviate the over-smoothing problem.

    \textbf{GIN}~\cite{xu2018gin} construct a straightforward architecture that is demonstrably the most expressive within the GNN class and matches the power of the Weisfeiler-Lehman graph isomorphism test.

    \textbf{MeanPooling}~\cite{xu2018gin} is a parameter-free pooling operation in graph neural networks. It generates a graph embedding by averaging the all node embeddings, encoding the complex graph into a unified vector representation.

    \textbf{TopKPooling}~\cite{gao2019_TopKPooling} introduces novel graph pooling and unpooling operations.
    The former adaptively selects nodes to form a smaller graph based on their scalar projection values on a trainable projection vector. 
    The latter, as the inverse operation, restores the graph to its original structure using the positional information of nodes selected in the corresponding pooling layer.

    \textbf{SAGPooling}~\cite{lee2019_SAGPooling} proposes a graph pooling approach based on self-attention. 
    By utilizing self-attention with graph convolution, this method considers both node features and graph topology. 

    \textbf{EdgePooling}~\cite{diehl2019_EdgePooling} proposes a graph pooling layer based on edge contraction. 
    This strategy learns a localized and sparse pooling transform to improve predictive performance. 
    It can be integrated into existing GNN architectures without adding any additional losses or regularization.

    \textbf{PANPooling}~\cite{ma2020_PANPooling} utilizes a convolution operation that considers every path linking the message sender and receiver, with learnable weights dependent on the path length, corresponding to the maximal entropy random walk. 
    This strategy offers a versatile framework adaptable to different graph data sizes and structures.

    Regarding (2), while some prevalent FL approaches (FedAvg~\cite{mcmahan2017fedavg}, FedProx~\cite{li2020fedprox}, Scaffold~\cite{karimireddy2020scaffold}, MOON~\cite{li2021moon}, and FedDC~\cite{gao2022feddc}) have demonstrated effectiveness in computer vision-based (CV-based) federated scenarios, recent FGL studies contend that they only achieve sub-optimal performance. 
    This weakness is attributed to their failure to incorporate topological information during the collaborative optimization process of FGL.
    Therefore, we only implement prevalent FL as baselines to assist future FGL developers in evaluating the effectiveness of their proposed methods. 
    
    Additionally, heterogeneity has been a persistent challenge in FL, encompassing multi-level heterogeneity arising from different local systems. 
    Specifically, this multi-level heterogeneity in terms of:
    (1) Data heterogeneity arises from variations in local data collection methods and data quality, leading to optimization challenges due to diverse multi-source data characterized by Non-iid data and domain shift;
    (2) Model heterogeneity stems from varying computational resource requirements, scalability needs, and predictive performance criteria across local systems, prompting the adoption of distinct local backbone models;
    (3) Communication heterogeneity is driven by variations in communication bandwidth among local devices, demanding the minimization of communication overhead. 
    Nonetheless, to enhance predictive performance in collaborative training, FL algorithms frequently necessitate increased information sharing, either at the client level or between clients and servers. 
    
    Building upon this foundation, to provide effective baselines, we have particularly implemented prototype-based methods (FedProto~\cite{tan2022fedproto}, FedNH~\cite{dai2023fednh}, FedTGP~\cite{zhang2024fedtgp}). 
    This has sparked a research trend in recent CV-based FL, as this technology can comprehensively address the aforementioned multi-level heterogeneity challenges.
    The prevalent graph-independent FL baseline details implemented in our proposed OpenFGL are listed below:

    \textbf{FedAvg}~\cite{mcmahan2017fedavg} serves as a foundational method in FL, enabling decentralized model training across diverse devices while preserving data privacy. 
    Initiated by a central server that distributes a global model, clients independently execute local updates.

    \textbf{FedProx}~\cite{li2020fedprox} allows for variable amounts of work to be performed locally across devices, and relies on a proximal term in model align loss to help stabilize the method.
    Theoretically, it offers convergence guarantees under conditions of non-identical data distributions and variable device workloads. 

    \textbf{Scaffold}~\cite{karimireddy2020scaffold} employs control variates to mitigate client-drift in FL. 
    Demonstrating significant reductions in communication rounds, Scaffold is resilient to data heterogeneity and client sampling. 

    \textbf{MOON}~\cite{li2021moon} is a model-contrastive FL framework that enhances local training by leveraging model representation similarities through contrastive learning at the model level.

    \textbf{FedDC}~\cite{gao2022feddc} is a novel FL algorithm that corrects local drift through lightweight modifications. 
    Each client tracks the deviation between local and global model parameters using an auxiliary variable, enhancing parameter-level consistency.

    \textbf{FedProto}~\cite{tan2022fedproto} is the first federated prototype learning framework for FL heterogeneity. 
    Instead of exchanging gradients, clients and the server share abstract class prototypes. 
    FedProto aggregates local prototypes and distributes global ones back to clients to regularize local model training.

    \textbf{FedNH}~\cite{dai2023fednh} addresses class imbalance by enhancing both personalization and generalization of local models. 
    FedNH distributes class prototypes uniformly in the latent space, infusing class semantics to prevent prototype collapse and enhance model performance. 
    This dual approach improves local models.
    
    \textbf{FedTGP}~\cite{zhang2024fedtgp} unlikes conventional methods that aggregate prototypes via weighted averaging, FedTGP uses adaptive contrastive Learning to train global prototypes on the server, enhancing prototype separability and preserving semantic integrity.

    As for (3), it is the core of OpenFGL as a comprehensive benchmark.
    To provide future FGL researchers with a comprehensive testing library and a user-friendly development framework, we conducted a thorough review of recent FGL studies, encompassing both the Graph-FL (GCFL+~\cite{xie2021gcfl}, FedStar~\cite{tan2023fedstar}) and Subgraph-FL (FedSage+~\cite{zhang2021fedsage}, Fed-PUB~\cite{baek2022fedpub}, FedGTA~\cite{li2023fedgta}, 
    FGSSL~\cite{huang2023fgssl},
    FedGL~\cite{chen2024fedgl}, AdaFGL~\cite{li2024adafgl},
    FGGP~\cite{wan2024fggp}, FedTAD~\cite{zhu2024fedtad}, FedDEP~\cite{zhang2024feddep}) scenarios, and comprehensively integrated them.
    The FGL baseline details implemented in our proposed OpenFGL are listed below:

    \textbf{GCFL+}~\cite{xie2021gcfl} dynamically clusters local systems using GNN gradient patterns to reduce structural and feature heterogeneity, particularly in the Graph-FL scenarios.
    Addressing the issue of fluctuating gradients, they enhance GCFL with a gradient sequence-based clustering mechanism using dynamic time warping, thereby improving clustering quality and theoretical robustness.

    \textbf{FedStar}~\cite{tan2023fedstar} shares structural embeddings across clients using an independent structure encoder. 
    This design allows FedStar to capture domain-invariant structural information while enabling personalized feature learning, thereby avoiding feature misalignment and enhancing inter-graph learning efficacy.

    \textbf{FedSage+}~\cite{zhang2021fedsage} integrates node features, link structures, and labels by employing a GraphSage model in conjunction with FedAvg to facilitate federated learning over local subgraphs. To further improve performance, FedSage+ introduces a generator to address missing links within the graph structure, thereby enhancing the model's robustness against incomplete data and ensuring a more comprehensive representation of graph relationships. This approach ultimately strengthens the ability of the model to generalize effectively across different clients in a federated learning setting.

    \textbf{Fed-PUB}~\cite{baek2022fedpub} is a novel framework for personalized subgraph FL that enhances local GNNs interdependently rather than forming a single global model. 
    Fed-PUB computes similarities between local GNNs using functional embeddings derived from random graph inputs, facilitating weighted averaging for server-side aggregation. 
    Additionally, it employs a personalized sparse mask at each client to selectively update subgraph-relevant parameters.

    \textbf{FedGTA}~\cite{li2023fedgta} innovatively merges large-scale graph learning with FGL.
    Clients encode topology and node attributes, compute local smoothing confidence and mixed moments of neighbor features, and then upload these to the server. 
    The server uses this data to perform personalized model aggregation, utilizing local smoothing confidence as weights for effective integration.

    \textbf{FGSSL}~\cite{huang2023fgssl} handles local client distortion in FL by focusing on node-level semantics and graph-level structures via the well-designed contrastive loss functions. 
    They enhance node discrimination by aligning local nodes with their global counterparts of the same class and distancing them from different classes. 
    Additionally, FGSSL transforms adjacency relationships into similarity distributions, using the global model to distill relational knowledge into local models, preserving both structure and discriminability.
    
    \textbf{FedGL}~\cite{chen2024fedgl} identifies global self-supervision information, which is then utilized to enhance prediction accuracy.
    Specifically, FedGL involves uploading prediction outcomes and node embeddings to the server to derive global pseudo labels and a global pseudo graph. 
    These global insights are distributed to each client, augmenting training labels and refining graph structures, consequently enhancing the performance of local models.

    \textbf{AdaFGL}~\cite{li2024adafgl} introduces a two-step personalized approach. In the first step, multi-client models are aggregated into a federated knowledge extractor at the server during the final communication round. In the second step, each client conducts personalized training using its local subgraph along with the federated knowledge extractor, allowing for more tailored model adaptation to local data characteristics while benefiting from collective knowledge.

    \textbf{FGGP}~\cite{wan2024fggp} divides the global model into two tiers linked by prototypes.
    At the classifier level, FGGP replaces traditional classifiers with clustered prototypes to enhance class discrimination and multi-domain prediction accuracy. 
    Meanwhile, at the feature extractor level, FGGP leverages contrastive learning to imbue prototypes with global knowledge, thereby improving model generalization.

    \textbf{FedTAD}~\cite{zhu2024fedtad} initially computes topology-aware node embeddings to evaluate the reliability of class-wise knowledge, transmitting this information to the server. 
    Guided by the class-wise knowledge reliability, FedTAD on the server side conducts data-free knowledge distillation to transfer reliable knowledge from local models across multiple clients to the global model.

    \textbf{FedDEP}~\cite{zhang2024feddep} leverages GNN embeddings for deep neighbor generation based on the FedSage+, employing efficient pseudo-FL for neighbor generation through embedding prototyping, and ensuring privacy protection via noiseless edge local DP. 
    Meanwhile, it utilizes prototype representation technologies to further reduce communication costs.

\vspace{0.3cm}
\subsection{Metric Description}
\label{appendix: Metric Description}
    Given the diverse downstream tasks in the FGL scenarios, we implement the following evaluation metrics tailored for regression (MSE, RMSE), classification (Accuracy, Precision, Recall, F1), prediction (AUC-ROC, AP), and clustering (Clustering-accuracy, NMI, ARI) tasks.
    Notably, in link prediction tasks within graph machine learning, AUC-ROC and AP are preferred over accuracy because they better handle the typical class imbalance, where non-existent links far outnumber actual links. 
    Accuracy can be misleading, as a model might achieve high accuracy by simply predicting the majority class.
    In contrast, AUC-ROC and AP focus on the model's ability to correctly rank positive links higher than negative ones, providing a more reliable evaluation of performance in scenarios where correct identification of the minority class (actual links) is crucial.

\textbf{Mean Squared Error}
    is a prevalent metric in graph regression tasks. 
    It measures the average squared deviation between predicted values and actual ground truth across the entire dataset. 
    Lower MSE values signify superior model performance, indicating a tighter alignment of predicted outcomes with observed results.

\textbf{Root Mean Squared Error} 
    is an extension of MSE and offers a more interpretable measure by taking the square root of the average squared differences between predicted and actual values. 
    Like MSE, lower RMSE denotes better alignment.

\textbf{Accuracy} 
    stands as a foundational metric in classification tasks, quantifying the ratio of correctly classified instances to the total instances in a dataset.
    It offers a clear indication of a graph learning model's overall predictive capability. 
    Higher accuracy values reflect a better alignment between predicted and actual class labels, demonstrating the model's effectiveness.

\textbf{Precision} 
    focuses on positive predictions. 
    Unlike Accuracy, it emphasizes the correctness of positive predictions by measuring the ratio of correctly predicted positive samples to all predicted positive samples. 
    This aspect becomes particularly critical when dealing with imbalanced datasets.

\textbf{Recall}
    measures a model's ability to capture all positive instances.
    Unlike Precision, it emphasizes correctly identified positive samples and overall actual positives. 
    This metric is vital in scenarios where missing positives have critical implications. 

\textbf{F1 Score} 
    represents the harmonic mean of Precision and Recall. 
    This metric provides a balanced assessment of a model's performance by considering both the precision of positive predictions and the model's ability to capture all positive instances. F1 Score is particularly valuable in scenarios where achieving high precision and recall are equally important. 

    \textbf{Area Under the Receiver Operating Characteristic curve} 
    quantifies the performance of a model in distinguishing between positive and negative links. 
    It provides a comprehensive measure of a model's ability to rank positive links higher than negative ones. 
    The high AUC-ROC indicates that the model effectively discriminates between positive and negative links.

    \textbf{Average Precision} measures the quality of a model's ranked list of positive links by calculating the average precision at each relevant position. 
    Unlike AUC-ROC, AP focuses solely on the precision-recall curve, providing a more detailed assessment of a model's performance, especially in imbalanced datasets where positive links are rare. 
    The high AP indicates that the model effectively ranks positive links higher than negative ones.

    \textbf{Clustering-accuracy} measures the agreement between the cluster assignments produced by a clustering algorithm and a ground truth clustering. 
    This metric differs from traditional accuracy metrics used in node classification, as it evaluates the overall coherence of cluster assignments rather than individual node labels. 
    A higher clustering accuracy indicates a better alignment between the identified clusters and the true underlying structure of the graph, thus reflecting the effectiveness of the clustering algorithm in uncovering meaningful communities or groups of nodes.

    \textbf{Normalized Mutual Information} quantifies the similarity between predicted clusters and ground truth by measuring the mutual information while normalizing for cluster size imbalances. 
    NMI ranges from 0 to 1, where higher values indicate better agreement between the predicted clusters and ground truth. 
    This metric is particularly valuable in scenarios where accurately identifying community structures or functional groups within a graph is critical.

    \textbf{Adjusted Rand Index} quantifies the similarity between predicted clusters and ground truth while considering the chance-corrected agreement. 
    ARI ranges from -1 to 1, where values closer to 1 indicate better agreement between the predicted clusters and ground truth than random clustering. 
    
\vspace{0.4cm}

\subsection{Robustness Simulation Description}
\label{appendix: Robustness Simulation Description}
    Given the practical applications driving FGL studies, the pivotal goal of various FGL approaches should be their effective deployment in real-world industrial scenarios. 
    Consequently, conducting a thorough evaluation of the robustness of existing methods becomes essential.
    In our proposal, besides exploring the generalization of current methods across various federated data simulation settings as discussed in Sec.~\ref{sec:benchmark_scenario}, we draw insights from common business challenges encountered in industrial scenarios.
    Specifically, we additionally integrate the following experimental setups to provide a comprehensive evaluation for industrial research projects from a robustness perspective.

\textbf{Feature Noise.}
    In real-world FGL applications, such as healthcare and finance, distributed privacy data is independently collected by local agents, leading to variations in data collection methods, processing techniques, and data sources. 
    These variations contribute to node feature noise, a common issue in practical settings.
    For example, in healthcare, differences in medical equipment, patient demographics, and data entry practices across hospitals can result in inconsistent patient data, introducing noise into the node features representing these data points. 
    In finance, variations in transactional systems, data aggregation methods, and processing protocols across institutions can lead to discrepancies in financial data, further contributing to node feature noise. 
    To accurately evaluate the robustness of existing FGL methods under such conditions, we simulate this scenario by introducing Gaussian or Laplacian noise into the node features of data samples within each client.
    Notably, Gaussian noise reflects natural data collection fluctuations, while Laplacian noise captures more pronounced deviations.

\textbf{Edge Noise.}
    In our implementation, we introduce two edge perturbation approaches: heterophilous noise (Subgraph-FL) and meta noise (Subgraph-FL and Graph-FL).
    For heterophilous noise, we randomly select non-connected node pairs for heterophilous perturbations based on their labels.
    This approach is motivated by recent studies~\cite{ma2021hete_gnn_survey1, luan2022hete_gnn_survey2, zheng2022hete_gnn_survey3}, which indicate that most GNNs struggle with heterophilous topology.
    Despite some GNN designs mitigating this issue, it has been rarely addressed in most FGL studies. 
    Moreover, heterophily is prevalent in the real world despite homophilous topology presumably dominating in default.
    As for meta noise, generated by Metattack~\cite{zugner_adversarial_2019_metaattck}, we budget the attack as 0.2 for each local dataset. 
    This approach shares the motivation of heterophilous noise but represents a more generalized and sophisticated perturbation method for both two FGL scenarios, which achieves optimal noise injection mechanisms through learnable means.
    In real-world applications, such as social networks, varying user interaction patterns across platforms can lead to inaccurate connections between nodes. 
    These discrepancies create noise edges, reflecting the inherent challenges of decentralized data environments.

\textbf{Label Noise.}
    Due to the diversity in data collection methods and the quality of local data sources, label noise is prevalent in crowd-sourcing scenarios. 
    In this scenario, the update process of the local model is inevitably affected, resulting in perturbed models. 
    When these model weights are uploaded to the server for multi-client collaboration, the naive federated paradigm suffers from global knowledge confusion, thereby significantly impacting the initialization of local models for the next round. 
    To simulate this setting, we introduce a novel perspective for evaluating algorithm robustness by randomly perturbing the true labels of training set samples according to a certain proportion.
    For instance, in crowd-sourcing, workers from diverse backgrounds may label the same data differently, leading to inconsistencies.
    These factors contribute to prevalent label noise, which disrupts the local model training process, resulting in perturbed models. 
    When these models are aggregated in the federated collaboration, the naive model aggregation mechanism struggles with global knowledge confusion, ultimately affecting the initialization of local models in the next training round. 
    To simulate this real-world challenge, we introduce a novel approach to evaluate algorithm robustness by randomly perturbing the true labels of training set samples according to a specified proportion.

\textbf{Feature/Edge/Label Sparsity.}
    In the current data explosion era, gathering substantial volumes of high-quality data can incur significant economic costs. Additionally, the laborious annotation requests both substantial manual labor and computational resources and leads to the prevalence of sparse data. 
    In graph-structured data, this sparsity challenge often manifests in missing attributes in feature dimensions, sparse graphs, and the well-known label sparsity issue. 
    To integrate the aforementioned scenarios into our proposed OpenFGL framework and evaluate the robustness of existing FGL studies from an industrial application perspective, we provide the following implementation details:
    In the feature sparsity setting, we assume that the feature representation of unlabeled nodes is partially missing. 
    To simulate edge sparsity, we randomly remove edges from subgraphs, providing a more challenging but realistic scenario.
    For label sparsity, we change the ratio of labeled nodes.

\textbf{Client Active Fraction/Client Sparsity.}
    In practical FGL scenarios, it is necessary to select a subset of clients to participate in each round to reduce communication costs or unavoidable device dropouts.
    However, the reduction in the number of clients participating in collaborative training during each communication round may lead to the global model deviating from the global optima. 
    In such a setting of \textit{client sparsity}, it is crucial to test whether FGL algorithms have the capability to accurately locate global optima through sparse data distribution. 
    Notably, client sparsity is a unique yet highly significant perspective for evaluating algorithm robustness in federated distributed scenarios.

    Based on the above noise and sparsity setting, we can evaluate the resilience and performance of FGL studies under conditions of data perturbation and practical scenarios, offering insights into their robustness in real-world applications where data and deployment environment may be imperfect.

\vspace{0.3cm}
\subsection{Effectiveness Evaluation Strategies}
\label{appendix: Effectiveness Evaluation Strategies}
    During our investigation, we observer a lack of descriptions for evaluating the effectiveness of existing FGL studies. 
    Therefore, in this section, we aim to present structured criteria from both data and model perspectives.
    These criteria aim to standardize the effectiveness evaluation for future FGL studies and support the experimental settings of this paper.

    To begin with, due to privacy regulations and prohibitively high manual costs, similar to FL in the computer vision domain, most existing FGL studies adopt a data partition strategy based on global data (i.e., benchmark datasets under centralized evaluation) to simulate distributed scenarios. 
    Based on this, we allocate the partitioned global data as multiple sets of local private data to different clients.
    Therefore, with the aforementioned data-driven experimental settings, we gain access to both implicit global data and local data from each client, providing an aspect for algorithm evaluation.
    Subsequently, in federated collaborative training, most FGL algorithms entail local training on private data at each client and model aggregation at the server side.
    This procedure gives rise to two perspectives for evaluating the models produced by the algorithms: 
    (1) server-side global model, typically transmitted from local models to the server and refined through well-designed server-side model aggregation or update mechanisms. 
    Since it integrates most local models from the current communication round, we refer to it as the global model.
    (2) client-side local model, mainly updated by local private data, which, in comparison to the global model, emphasizes fitting local data more closely.
    It is often emphasized by personalized algorithms, as they focus on the local training frameworks.
    To this end, considering both data and model perspectives, we acquire global data and local data, along with the global model and local model, respectively.
    Combining these aspects results in four effectiveness evaluation strategies in FGL, which reflect the diverse business requirements of practical industrial scenarios. 
    In a nutshell, we derive the following four evaluation criteria:

    (1) Global Model on Global Data:
    This primarily aims to verify the generalization of current FGL algorithms by evaluating the performance of the server-side collaborative model on a broader empirical data domain (i.e., global data). 
    This evaluation is conducted more for experimental analysis from a generalization perspective during the research process.

    (2) Global Model on Local Data: 
    This evaluation is designed to facilitate multi-client collaborative training facilitated by a trusted server, leveraging diverse knowledge to enhance the robustness of the server-side collaborative model without direct data sharing.
    In practical applications, multiple clients undergo standardized training and evaluation under the supervision of a trusted server.

    (3) Local Model on Global Data: 
    Similar to (1), this evaluation is aimed at empirically analyzing personalized FGL algorithms from a generalization perspective. 
    Considering that personalized algorithms highlight the distinct neural architectures or learning mechanisms of individual local clients, essential generalization analysis is conducted to determine whether the current algorithm can produce unbiased predictions and mitigate over-fitting issues. 
    This analysis involves evaluating the performance of client-side personalized models on global data.

    (4) Local Model on Local Data: 
    This evaluation represents the most application-driven evaluation strategy among the above strategies, as it aligns with the practical requirements of FL.
    In FL, multiple clients collaborate to enhance their local scenarios with more robust personalized models while addressing privacy concerns. 
    Consequently, in our comprehensive benchmark incorporating multiple algorithms, we default to this effective evaluation strategy. 
    This approach, driven by practical applications, serves to verify the real-world deployment feasibility of current algorithms.
    
\subsection{Experiment Environment}
\label{appendix: Experiment Environment}
    The experiments are conducted on the machine with Intel(R) Xeon(R) Gold 6240 CPU @ 2.60GHz, and NVIDIA A100 80GB PCIe and CUDA 12.2. The operating system is Ubuntu 18.04.6 with 216GB memory.
    As for software versions in the environment, we use Python 3.9 and Pytorch 1.11.0.

\vspace{0.2cm}

\subsection{Hyperparameter Settings}
\label{appendix: Hyperparameter Settings}

\textbf{General Experimental Settings.} 
    For Graph-FL, the learning rate is typically set to  \(1 \times 10^{-3}\), with each client performing 1 epoch per communication round and a batch size of 128.
    In Subgraph-FL, the learning rate is raised to \(1 \times 10^{-2}\), and the local epochs are extended to 3. 
    This adjustment accommodates the typically larger scale of node samples in Subgraph-FL, necessitating a larger learning rate and more local iterations to facilitate model convergence.
    Additionally, for both scenarios, we standardize certain parameters. 
    The weight decay is set to \(5 \times 10^{-4}\), the number of communication rounds is set to 100, the dropout rate is set to 0.5, and optimization is conducted using the Adam~\cite{kingma2014adam} optimizer. 
    To evaluate the robustness of our results under varying initial conditions, we eliminate the use of fixed random seeds. 
    All experiments are repeated three times to report the mean and variance of the respective metrics for unbiased predictive performance.

\textbf{Personalized Baseline Settings.} 
    We perform extensive hyperparameter tuning to ensure a comprehensive and unbiased evaluation of these FGL methods using the Optuna framework~\cite{akiba2019optuna}. 
    The hyperparameter search spaces for all baselines are available in our GitHub repository.
    For detailed explanations of these hyperparameters, please refer to their original papers.

    Regarding graph-specific data simulation strategies in Table~\ref{tab: simulation}, to enhance readability and avoid complex figures or tables, we default to using 10-client label Dirichlet (i.e., Label Distribution Skew and $\alpha=1$) and Metis partitioning (i.e., Metis-based Community Split) separately for the Graph-FL and Subgraph-FL scenarios. 
    The former is inspired by data Non-iid simulation in CV~\cite{li2020fedprox, karimireddy2020scaffold, li2021moon}, while the latter is inspired by prevalent data simulation strategies in current FGL studies~\cite{baek2022fedpub, li2023fedgta, li2024adafgl}. 
    Experimental evaluations of the generalization of existing methods across different data simulation scenarios can be found in Sec.~\ref{sec: Robustness Analysis}.
    Furthermore, in the selection of local backbone models for graph learning, we choose prevalent GIN and GCN models, applied to Graph-FL and Subgraph-FL, respectively.
    Notably, we experiment with multiple datasets and baselines in separate modules and use graph/node classification to report experimental results to further avoid complex charts, making the results more reader-friendly.
    
\vspace{0.1cm}

\subsection{DP-based Privacy Persevere}
\label{appendix: DP-based Privacy Persevere}
\subsubsection{Preliminaries on Differential Privacy}
\label{appendix: DP-preliminaries}
    DP~\cite{dwork2006calibrating} has become the dominant model for the protection of individual privacy from powerful and realistic adversaries. 
    Informally, it requires that the output of a differentially private query is not dramatically affected by the inclusion or exclusion of any particular individual's data in the input. 
    This means that even if an attacker can access all but one individual's data, they cannot determine whether it was included in the computation. The formal definition of DP is as follows:
\begin{definition}[Differential Privacy~\cite{dwork2006calibrating}]\label{def:DP}
    Let $ D $ and $ D^{\prime} $ be two adjacent datasets that differ in at most one entry. A randomized algorithm $ \mathcal{A} $ satisfies $(\epsilon,\delta)$-differential privacy if for all $O \subseteq Range(\mathcal{A})$:
    \begin{align}
        [\mathcal{A}(D)\in O] \leq e^{\epsilon} \cdot [\mathcal{A}(D^{\prime})\in O]+\delta.
    \end{align}
\end{definition}
    The privacy budget $\epsilon$ controls the trade-off between the level of privacy protection and utility: a lower $\epsilon$ indicates stricter privacy preservation but leads to lower utility. 
    The parameter $\delta$ represents the maximum permissible failure probability and is usually chosen to be much smaller than the inverse of the number of data records.

    The formal definition of DP revolves around the concept of adjacency between datasets. 
    When data are represented as a graph, two notions of adjacency are defined: \textit{edge}-level and \textit{node}-level adjacency. 
    Edge-level adjacency occurs when two graphs differ by just one edge, while node-level adjacency involves a difference in an entire node and its associated features, labels, and connections. 
    Therefore, an algorithm $\mathcal{A}$ provides edge-level (or node-level) $(\epsilon,\delta)$-DP if for any two edge-level (or node-level) adjacent graph datasets $G$ and $G^{\prime}$ and any possible outputs $O\subseteq Range(\mathcal{A})$, the inequality $[\mathcal{A}(G)\in O] \leq e^{\epsilon} \cdot [\mathcal{A}(G^{\prime})\in O]+\delta$ holds. 
    Edge-level DP focuses on the protection of edge privacy, while node-level DP aims to protect the privacy of nodes and their connections. 
    In consequence, the node-level DP can provide more robust privacy protection.
    
    In this paper, we use an alternative definition of DP, called \textit{R\'enyi Differential Privacy} (RDP)~\cite{mironov2017renyi}, since it allows for tighter composition of DP across multiple steps.

    \begin{definition}[R\'enyi Differential Privacy~\cite{mironov2017renyi}]
        An randomized algorithm $ \mathcal{A} $ is said to be $ (\alpha, \gamma) $-RDP, if, for every pair of adjacent datasets $ G $ and $ G^{\prime} $, we have
        \begin{align}
            D_\alpha(\mathcal{A}(G)\|\mathcal{A}(G^{\prime}))\leq \gamma,
        \end{align}
        where $ D_\alpha(P\|Q) $ is the R\'enyi divergence of order $ \alpha $ between probability distributions $ P $ and $ Q $ defined as:
        \begin{align}
            D_\alpha(P\|Q)=\frac{1}{\alpha-1}\log\mathbb{E}_{x\sim Q}\left[\frac{P(x)}{Q(x)}\right]^\alpha.
        \end{align}
    \end{definition}
    The concept of RDP is closely related to the standard $ (\epsilon,\delta) $-DP. 
    According to \cite{mironov2017renyi}, any mechanism that achieves $ (\alpha, \gamma) $-RDP also fulfills $ (\gamma+\log(1/\delta)/(\alpha-1),\delta) $-DP for any $ \delta\in(0,1) $. 
    A basic mechanism to achieve RDP is the Gaussian mechanism. Specifically, we inject Gaussian noise into the algorithm's output for privacy protection. 
    And the noise follows the Gaussian distribution $ N(0, \alpha\Delta_{2}^{2}/2\gamma) $, where $ \Delta_{2} $ represents the $ \ell_{2} $-sensitivity.

    \begin{definition}[$ \ell_{2} $-Sensitivity]\label{def:GS}
        Given a function $ f $: $ \mathcal{G} \to \mathbb{R}^{d} $, the $ \ell_{2} $-sensitivity of $ f $ is defined as
        \begin{align}
            \Delta_{2} = \max_{G, G^{\prime}}||f(G)-f(G^{\prime})||_{2},
        \end{align}
        where $ G $ and $ G^{\prime} $ are adjacent datasets and $ ||\cdot||_{2} $ is the $ \ell_{2} $ norm.
    \end{definition}

\vspace{0.1cm}
\subsubsection{Privacy-Preserving Techniques in OpenFGL}
\label{appendix: detail-privacy-techniques} 
    In federated collaboration, each client uploads model gradients to the server. 
    However, the gradients computed directly from raw data are susceptible to inference attacks~\cite{nasr2019comprehensive, luo2021feature} and reconstruction attacks~\cite{zhu2019deep, geiping2020inverting}, which can lead to privacy breaches. 
    OpenFGL implements basic privacy-preserving techniques that satisfy node-level RDP to protect individual privacy when model gradients are uploaded. 
    The core idea is that each client perturbs the gradients via the Gaussian mechanism and then sends a perturbed version to the server.

    Specifically, each client receives initial parameters from the server. 
    Then, similar to the standard mini-batch SGD technique, each client samples a subset $ \mathcal{S} $ that consists of $ k $ samples selected uniformly at random from the training set. 
    Clients can compute the gradients via forward and backward propagation within this mini-batch. 
    Given that there is no a priori constraint on the size of the model gradients, we employ the clipping operator $ Clip_{C} $ to handle the gradient of each sample $ \mathbf{w}_{i} $: $ Clip_{C}(\mathbf{w}_{i})=\mathbf{w}_{i}\cdot\min(1, C/||\mathbf{w}_{i}||_{2}) $, where $ C $ is the clipping threshold.
    However, in the context of GNNs, all direct and multi-hop neighbors participate in the calculation of gradients for each node via recursive layer-wise message passing~\cite{hamilton2017inductive}. 
    At each layer, the representation of each node is derived not solely from its features but also from the features of adjacent nodes. 
    Therefore, each per-sample gradient term can be influenced by private data from multiple nodes~\cite{daigavane2021node}. 
    This means the sensitivity of the gradient due to the presence or absence of a node can be extremely high due to the node itself and its neighbors, which makes standard DP-SGD-based methods~\cite{abadi2016deep} infeasible, resulting in either high privacy cost or poor utility due to the considerable required DP noise. 
    In this paper, we restrict the number of graph convolutional layers and study models with only one GNN layer. 
    Under this limitation, the sensitivity of $ 1 $-Layer GNN as follows:
    \begin{lemma}[Node-Level Sensitivity of the $ 1 $-Layer GNN~\cite{daigavane2021node}]
        For any node $ v_{i} $, let $ \mathbf{y}_{i} $ represent the ground truth and $ \mathbf{\tilde{y}}_{i} $ the prediction from a $ 1 $-layer GNN. Consider the loss function $ \mathcal{L} $ of the form: $ \mathcal{L}(G,\mathbf{\Theta})=\sum_{v_{i}\in \mathcal{V}}\ell(\tilde{\mathbf{y}}_{i};\mathbf{y}_{i}) $. The $ \ell_{2} $-sensitivity of the aggregated gradient, $ \mathbf{w}_G=\sum_{v_{i}\in\mathcal{S}}Clip_C(\nabla_{\mathbf{\Theta}}\ell(\tilde{\mathbf{y}}_i;\mathbf{y}_i)) $, is determined by the equation:
        \begin{align}
            \Delta_{2}(\mathbf{w}_G)=2(d_{max}+1)C,
        \end{align}
        where $ d_{max} $ denotes the maximum degree of the graph $ G $.
    \end{lemma}
    
    To achieve a better trade-off between privacy and utility, we also utilize the privacy amplification technique~\cite{daigavane2021node}, which is implemented by sampling. 
    Next, we sample noise from a Gaussian distribution $ N(0, \alpha\Delta_{2}^{2}(\mathbf{w}_G)/2\gamma) $ and add the noise to the aggregated gradient over the mini-batch. 
    Finally, the perturbed gradients are sent to the server for model aggregation. 
    The server aggregates the perturbed gradients from all clients and updates the global model parameters. 
    The process is repeated until the model converges.

    However, when a smaller privacy budget is allocated, there is a significant degradation in performance, as illustrated in Table~\ref{tab: dp privacy preserve}. 
    This decline is primarily attributed to the unique characteristics of the GNN model, as previously mentioned. 
    This uniqueness results in a high sensitivity of the model gradient and affects the privacy amplification techniques. 
    Consequently, substantial noise is introduced into the gradients during training, which markedly disrupts the learning process.
    This indicates that training graph neural networks in federated scenarios still faces a tough test. 
    Therefore, it is crucial to develop new privacy-preserving techniques that can effectively protect privacy while maintaining performance.

    For Graph-FL, where each individual sample represents an entire graph, the application of DP-SGD~\cite{abadi2016deep} can be directly extended to these tasks.
    This approach not only preserves privacy but also optimizes the model effectively, making it particularly suitable for scenarios where maintaining the integrity and confidentiality of the entire graph as a cohesive data unit is crucial. 
    This method is ideal for ensuring that the learning process respects the privacy constraints inherent in sensitive data environments.

\subsubsection{Technique Details of FedDEP}
\label{appendix: detail-feddep}
    In addition, we have implemented the FedDEP~\cite{zhang2024deep} algorithm within the OpenFGL framework. 
    FedDEP achieves noise-free edge-level DP by employing random sampling, ensuring robust privacy preservation without compromising data integrity. 
    To tackle the issue of cross-subgraph missing neighbors, FedDEP incorporates an advanced deep neighbor generation module known as DGen, which enhances the model's capability to generate and integrate missing neighbors effectively.
    To further optimize computational efficiency within each client, local GNN embeddings are clustered to create sets of missing neighbor prototypes. 
    Moreover, to reduce inter-client communication overhead, FedDEP introduces a pseudo-federated learning approach, where these prototype embeddings are shared across the system before the training of DGen, thereby streamlining the collaborative learning process while maintaining model accuracy and privacy. 

    To protect privacy, FedDEP adapts noise-free differential privacy, originally developed for general domains~\cite{sun2020federated}, to edge-level local differential privacy. 
    Specifically, it implements two random sampling strategies during the FedDEP model training: 
    (i) random neighborhood sampling within each graph convolutional layer, and 
    (ii) random sampling of the generated deep neighborhoods by a Bernoulli sampler. 
    These strategies collectively aim to obscure the individual contributions of nodes, which ensures privacy guarantees with the smallest possible impact on the accuracy of the model. 
    However, the FedDEP algorithm is designed for edge-level DP, which only protects the privacy of edge and not node features or node embeddings. 
    This limitation may lead to potential privacy breaches when the node features contain sensitive information.

\vspace{0.2cm}
\subsubsection{Future Research Directions}
\label{appendix: future-work-privacy}
    This section introduces several promising research directions in DP-based FGL. 
    Currently, there are few privacy-preserving algorithms specifically tailored for FGL. 
    Given the distinctive properties of graph data, traditional privacy mechanisms commonly employed in FL fail to seamlessly extend to this domain. 
    Therefore, it is crucial to develop new mechanisms for FGL that should consider the unique characteristics of GNNs to achieve an optimal trade-off between privacy protection and model utility. 
    Furthermore, the convergence analysis of differentially private FGL algorithms is still an open research problem. 
    A tight convergence upper bound not only theoretically assures rapid convergence, but also facilitates an empirical examination of how various hyperparameters influence convergence rates. 
    Such insights are pivotal for fine-tuning parameters or the development of innovative optimization strategies~\cite{li2024convergence}. 
    Existing analyses, such as those presented in~\cite{wei2020federated, li2022soteriafl}, do not consider GNN-specific processes like propagation and aggregation. 
    Consequently, there is an urgent need for advanced convergence analysis approaches suitable for DP-FGL. 
    Another important research direction is to evaluate the performance of differentially private FGL algorithms against malicious attacks, such as inference attacks~\cite{nasr2019comprehensive, luo2021feature} and reconstruction attacks~\cite{zhu2019deep, geiping2020inverting}.
    This research will contribute significantly to the robustness and reliability of DP-based FL.

\subsection{Federated Heterogeneous Graph Learning}
\label{appendix: Federated Heterogeneous Graph Learning}
    Heterogeneous graphs (HGs), which are characterized by multiple types of nodes and relations, are widely encountered in real-world scenarios, including social networks and recommendation systems. 
    These graphs offer a more comprehensive representation of complex systems, encompassing diverse information and richer semantics compared to homogeneous graphs, making them particularly valuable for modeling and analysis in various practical applications.

    Therefore, HGs in federated settings present more complex scenarios and pose greater challenges for federated learning. 
    Beyond the typical feature heterogeneity, label heterogeneity, and structural heterogeneity encountered in homogeneous graphs within FGL, the diverse relationships inherent in HGs introduce an additional layer of complexity—relation heterogeneity among different clients~\cite{DBLP:conf/www/Xie0Y23}.
    This diversity in relational types across clients complicates the learning process, as it requires handling variations in how different clients structure and interpret these relationships. 
    Moreover, since meta-paths within HGs carry specific semantic information, differences in relation types lead to semantic heterogeneity among clients, further complicating the design of effective FGL algorithms~\cite{DBLP:conf/www/YanCWYDS24}. 
    Consequently, existing federated heterogeneous graph learning (FHGL) methods are often highly tailored to specific scenarios and needs, leading to a lack of standardized approaches for these complex learning environments. 
    This variability underscores the need for more generalized and adaptable FHGL frameworks that can effectively manage the diverse challenges posed by HGs in federated settings.

    In the following subsection, we provide a comprehensive summary of 6 FHGL models, categorizing them into two setups based on their application scenarios: \textit{Relation Type Sharing} and \textit{Relation Type Protection}. 
    For each algorithm, we outline the core ideas and detail their FL strategies, encoders, and client partitioning approaches, as presented in Table~\ref{Table_FHGL}. 
    Additionally, we introduce a basic FHGL framework within OpenFGL to assist users in efficiently conducting FL experiments on heterogeneous graphs, thereby facilitating more streamlined research and development in this area.

\begin{table*}[htbp]
  \centering
  \caption{Summary of federated heterogeneous graph learning algorithms. Asterisk (*) indicates that the model has been improved, and "None" indicates that it is not described in the original paper.}
    \begin{tabular}{rccccc}
    \toprule
    \multicolumn{1}{c}{\textbf{Client Setup}} & \textbf{Method} & \textbf{Basic FL} & \textbf{HGNN Encoder} & \textbf{Datasets} & \textbf{Graph Partitioning} \\
    \midrule
          & \textbf{FDRS} \cite{DBLP:conf/hpcc/LiLML23} & FedAvg \cite{mcmahan2017fedavg} & HGCN* & Epinions & None \\
\cmidrule{2-6}          & \multirow{3}[2]{*}{\textbf{FedAHE} \cite{DBLP:conf/ccis/WangLLL22}} & \multirow{3}[2]{*}{FedDWA} & \multirow{3}[2]{*}{HAN \cite{han}} & ACM   & \multirow{3}[2]{*}{None} \\
          &       &       &       & DBLP  &  \\
          &       &       &       & Aminer &  \\
\cmidrule{2-6}    \multicolumn{1}{c}{\textbf{Relation}} & \multirow{4}[2]{*}{\textbf{FedDA} \cite{DBLP:conf/icde/GuZBCZY23}} &       & \multirow{4}[2]{*}{Simple-HGN \cite{DBLP:conf/kdd/LvDLCFHZJDT21}} & DBLP  & \multirow{4}[2]{*}{Dominant} \\
    \multicolumn{1}{c}{\textbf{Type}} &       & Dynamic  &       & Amazon &  \\
    \multicolumn{1}{c}{\textbf{Sharing}} &       & Activation &       & LastFM &  \\
          &       &       &       & PubMed &  \\
\cmidrule{2-6}          & \multirow{4}[2]{*}{\textbf{FedHGNN} \cite{DBLP:conf/www/YanCWYDS24}} & \multirow{4}[2]{*}{FedAvg} & \multirow{4}[2]{*}{HAN} & ACM   & \multirow{4}[2]{*}{Ego Graph} \\
          &       &       &       & DBLP  &  \\
          &       &       &       & Yelp  &  \\
          &       &       &       & Douban &  \\
    \midrule
          & \multirow{3}[2]{*}{\textbf{FedHGN} \cite{DBLP:conf/ijcai/0004K23}} & \multirow{3}[2]{*}{FedAvg*} & \multirow{3}[2]{*}{RGCN* \cite{DBLP:conf/esws/SchlichtkrullKB18}} & AIFB  & Random Edges \\
          &       &       &       & MUTAG & Random Edge Types \\
    \multicolumn{1}{c}{\textbf{Relation}} &       &       &       & BGS   &  \\
\cmidrule{2-6}    \multicolumn{1}{c}{\textbf{Type}} & \multirow{4}[2]{*}{\textbf{FedLIT} \cite{DBLP:conf/www/Xie0Y23}} &       & \multirow{4}[2]{*}{RGCN*} & DBLP  & Distinct \\
    \multicolumn{1}{c}{\textbf{Protection}} &       & Dynamic &       & PubMed & Dominant \\
          &       & Clustering &       & NELL  & Balanced \\
          &       &       &       & MIMIC3 &  \\
    \bottomrule
    \end{tabular}%
  \label{Table_FHGL}%
\end{table*}%

\vspace{0.1cm}

\subsubsection{Relation Type Sharing}
    This FHGL setup typically defines global relation types on the server side, with each client constructing its local heterogeneous graph according to these predefined standards.
    The focus is primarily on addressing the challenges associated with federated training instability and performance degradation that arise due to relation heterogeneity across clients.
    
    As one of the pioneering FHGL models, FDRS~\cite{DBLP:conf/hpcc/LiLML23} addresses the cold start problem caused by sparse data within individual clients. 
    FDRS utilizes a two-level aggregation heterogeneous graph convolutional network (HGCN) within each client. 
    This approach facilitates message passing between nodes of the same type through object-level aggregation, while type-level aggregation updates information across different types of nodes. 
    After local training, each client uploads its model parameters to the server, where the FedAvg~\cite{mcmahan2017fedavg} algorithm aggregates and updates the parameters from all clients.
    Although FDRS does not implement a specific communication strategy tailored to FHGL, it effectively demonstrates the benefits of cross-client HGNNs. 
    Moreover, it underscores the necessity of learning heterogeneous relations from other clients to enhance and enrich the local information available to each client, thereby improving overall model performance.

    FedAHE~\cite{DBLP:conf/ccis/WangLLL22} underscores the significance of recognizing the diversity of meta-path instances in HGs, stressing that this diversity should be accounted for not only within individual clients but also across different clients.
    To address the semantic heterogeneity that arises among clients due to differences in meta-paths, the authors propose dynamic weighted aggregation of parameters (FedDWA). 
    This mechanism aims to harmonize the variations in meta-paths, thereby reducing semantic discrepancies across clients.
    During the training process, after each round of aggregation on the server, FedAHE evaluates whether the model version gap between clients exceeds a predefined threshold. 
    If the version gap surpasses this limit, the server initiates a synchronization process by distributing the latest model weights to all clients, ensuring that local models are updated and aligned with the most recent global model. 
    This approach helps to maintain consistency across the federation, thereby enhancing the overall robustness and effectiveness of the FL process in the presence of meta-path diversity.

    Compared to FGL algorithms, FedDA~\cite{DBLP:conf/icde/GuZBCZY23} analyzes the unique characteristics of FL on HGs: that is, only a small amount of client parameters need to be uploaded in each communication round to achieve rapid convergence. 
    Based on these findings, FedDA proposes a dynamic activation strategy, which achieves efficient training by dynamically selecting a subset of clients for each round of aggregation. 
    Furthermore, considering the complex relations in HGs, FedDA introduces the D-HGN model, which decouples the parameters of relation types to allow partial updates of relation parameters on the server side instead of the entire model parameters.
    
    FedHGNN~\cite{DBLP:conf/www/YanCWYDS24} identifies the semantic broken issue that may arise due to the incompleteness of HGs across different clients. 
    To mitigate this issue, FedHGNN proposes a semantic-preserving user interaction publishing algorithm, which captures cross-client semantic information by uploading a shared pattern to the server side. 
    Furthermore, to prevent the shared pattern from leaking the local client's privacy, FedHGNN introduces a two-stage perturbation mechanism to disturb the interactions within the local client and theoretically verifies that this strategy satisfies both semantic privacy and interaction privacy guided by semantics. 
    This strategy only requires one communication before the FL training and then utilizes HAN~\cite{han} for encoding within the client and FedAvg for cross-client communication training.

\subsubsection{Relation Type Protection}
    This FHGL setup assumes that the relation type of each client is private and protected, and the server cannot know the specific relation types of clients. 
    Therefore, related models usually adopt some heuristic manners to achieve personalized aggregation and updating of local model parameters without exposing the specific client-side edge relations.
    
    FedHGN~\cite{DBLP:conf/ijcai/0004K23} highlights that the complex relations in HGs make it challenging for clients to collect and maintain all types of relations. 
    To address this issue, they propose a schema-weight decoupling strategy, which involves performing basis decomposition on the weight matrix of locally specific relations to form relation-specific coefficients $\beta$ and globally shared basis decomposition. 
    Meanwhile, FedHGN updates $\beta$ by heuristically matching the minimum distance between relation-specific coefficients of different clients.
    
    FedLIT~\cite{DBLP:conf/www/Xie0Y23} is primarily designed for vertical FGL. 
    For example, in the same city, different institutions may have similar user samples, but due to institutional differences (such as hospitals and shopping malls), the relations between users are also different. 
    To address this, FedLIT proposes a dynamic latent link-type-aware clustered strategy. 
    This strategy clusters within clients based on edge-type embeddings to obtain local centroids for each relation, and then performs a secondary clustering of the local centroids on the server side to obtain global centroids. 
    Similarly, FedLIT employs heuristic methods to group each local centroid on the server side and aggregates and updates specific relation projection matrice.

\begin{table*}[htbp]
  \centering
  \caption{Statistics of heterogeneous graph datasets.}
    \begin{tabular}{cccccccc}
    \toprule
    \textbf{Dataset} & \textbf{\#Node} & \textbf{\#Edge} & \multicolumn{2}{c}{\textbf{\#Node Type}} & \multicolumn{2}{c}{\textbf{\#Relation}} & \textbf{\#Classes} \\
    \midrule
    \multirow{2}[2]{*}{\textbf{ACM}} & \multirow{2}[2]{*}{10,942} & \multirow{2}[2]{*}{547,872} & \# Author (A) & \# Term (T) & P$\rightleftharpoons$ A    & P$\rightleftharpoons$ T    & \multirow{2}[2]{*}{3} \\
          &       &       & \# Paper (P) & \# Subject (S) & P$\rightleftharpoons$ P    & P$\rightleftharpoons$ S    &  \\
    \midrule
    \multirow{2}[2]{*}{\textbf{DBLP4HGB}} & \multirow{2}[2]{*}{26,128} & \multirow{2}[2]{*}{239,566} & \# Author (A) & \# Term (T) & A$\rightleftharpoons$ P    & P$\rightleftharpoons$ T    & \multirow{2}[2]{*}{4} \\
          &       &       & \# Paper (P) & \# Venue (V) & P$\rightleftharpoons$ V    &       &  \\
    \midrule
    \multirow{2}[2]{*}{\textbf{DBLP4MGN}} & \multirow{2}[2]{*}{26,128} & \multirow{2}[2]{*}{296,563} & \# Author (A) & \# Term (T) & A$\rightleftharpoons$ P    & P$\rightleftharpoons$ T    & \multirow{2}[2]{*}{4} \\
          &       &       & \# Paper (P) & \# Conference (C) & P$\rightleftharpoons$ C    &       &  \\
    \midrule
    \multirow{2}[2]{*}{\textbf{IMDB}} & \multirow{2}[2]{*}{21,420} & \multirow{2}[2]{*}{86,624} & \# Movie (M) & \# Director (D) & \multirow{2}[2]{*}{M$\rightleftharpoons$ D} & \multirow{2}[2]{*}{M$\rightleftharpoons$ A} & \multirow{2}[2]{*}{3} \\
          &       &       & \# Actor (A) &       &       &       &  \\
    \bottomrule
    \end{tabular}%
  \label{Table_hg}%
  \vspace{0.08cm}
\end{table*}%

\begin{table*}[htbp]
  \centering
  \caption{The performances (\%) of federated heterogeneous graph learning on node classification.}
    \begin{tabular}{ccccccc}
    \toprule
    \multicolumn{2}{c}{Methods} &       & \multicolumn{4}{c}{Datasets} \\
\cmidrule{1-2}\cmidrule{4-7}    Basic FL & HGNN Encoder &       & ACM   & DBLP4HGB & DBLP4MGN & IMDB \\
    \midrule
    \multirow{3}[2]{*}{FedAvg \cite{mcmahan2017fedavg}} & HAN   &       & 86.36±1.40 & 78.32±4.17 & 77.19±4.26 & 50.39±0.98 \\
          & RGCN  &       & 89.74±0.24 & 80.86±0.59 & 80.39±0.47    & 60.42±0.42 \\
          & HGT   &       & 84.25±1.30 & 79.29±0.78 & 78.28±1.53    & 55.86±0.73 \\
    \midrule
    \multirow{3}[2]{*}{FedDC \cite{gao2022feddc}} & HAN   &       & 83.81±1.03 & 72.55±4.96 & 73.62±4.62    & 52.31±0.95 \\
          & RGCN  &       & 71.53±4.45 & 81.23±0.57 & 81.27±0.47    & 60.95±0.62 \\
          & HGT   &       & 85.22±2.16 & 79.61±0.78 & 78.28±1.53   & 50.57±0.91 \\
    \midrule
    \multirow{3}[2]{*}{Moon \cite{li2021moon}} & HAN   &       & 86.03±1.46 & 77.02±3.91 & 77.19±3.99    & 51.30±1.11 \\
          & RGCN  &       & 89.50±0.51 & 81.24±0.42 & 81.42±0.38    & 60.99±0.72 \\
          & HGT   &       & 82.25±2.53 & 77.75±1.36 & 79.42±1.26     & 56.80±1.11 \\
    \midrule
    \multirow{3}[2]{*}{Scaffold \cite{karimireddy2020scaffold}} & HAN   &       & 86.66±1.62 & 77.66±4.18 & 78.02±4.12    & 55.54±0.60 \\
          & RGCN  &       & 89.87±0.32 & 81.07±0.65 & 80.86±0.76    & 59.02±0.07 \\
          & HGT   &       & 87.78±0.91 & 79.24±0.83 & 79.46±0.71 & 55.96±0.75 \\
    \bottomrule
    \end{tabular}%
    \vspace{0.08cm}
  \label{Table_fhgl_exp}%
\end{table*}%

\subsubsection{Basic FHGL on OpenFGL}
    Due to the inherent complexity of HGs, the aforementioned algorithms are designed only for specific scenarios rather than a general federated graph scenario. 
    To facilitate users in quickly conducting FGL tasks on HG datasets, we provide a basic FHGL model on OpenFGL for users to perform simulation experiments. 
    Specifically, OpenFGL offers two types of heterogeneous graph partitioning strategies: 
    (1) \textbf{Relation Type Sharing}: This strategy follows the traditional FL setup, where nodes of the target type are partitioned among multiple clients according to a Dirichlet distribution, and other types of nodes are extracted based on their relations with the target nodes. 
    In this way, the relations and node types in each client are the same, and these types are considered globally shared. 
    (2) \textbf{Relation Type Protection}: According to the Random Edge Types strategy provided by FedHGN~\cite{DBLP:conf/ijcai/0004K23}, different types of relations are randomly partitioned among different clients. 
    This method requires that the relation types in each client cannot be shared, necessitating the establishment of specific FL strategies to protect this privacy.

    For heterogeneous graph datasets, we used DBLP4HGB and ACM datasets provided by Simple-HGN~\cite{DBLP:conf/kdd/LvDLCFHZJDT21}, as well as the DBLP4MGN and IMDB datasets provided by MAGNN~\cite{magnn}. 
    The statistical information of the datasets is shown in Table~\ref{Table_hg}. 
    Here, we provide some experimental results, as shown in Table~\ref{Table_fhgl_exp}. 
    The heterogeneous subgraph partitioning strategy follows the relation type sharing strategy and is divided into 10 clients. 
    All experiments are repeated ten times, and the mean and standard deviation are reported.

    Although in FHGL, different HGNNs~\cite{DBLP:conf/esws/SchlichtkrullKB18, han, DBLP:conf/www/HuDWS20} and FL methods~\cite{mcmahan2017fedavg, gao2022feddc, li2021moon, karimireddy2020scaffold} can be combined to achieve the basic framework, it lacks distributed characteristics. 
    For example, in centralized learning, the attention mechanism is considered an effective method of identifying relations~\cite{DBLP:conf/kdd/LvDLCFHZJDT21}. 
    However, in Table~\ref{Table_fhgl_exp}, HAN and HGT based on different attention mechanisms perform worse in the FHGL scenario compared to the RGCN. 
    We speculate that this is because the attention mechanism requires additional parameters, making it more prone to over-fitting and stronger local biases. 
    Additionally, the information loss caused by meta-paths is further amplified in FGL, resulting in HAN achieving the worst performance.
    It is worth mentioning that in most cases, the performance of the same HGNN model shows almost no significant differences across different FL methods. 
    Therefore, for distributed HGs, designing more reasonable HGNNs seems to be a more effective approach.
    OpenFGL has user-friendly extensibility, allowing users to quickly experiment with their own HGNN and FHGL strategies.
    Given the practical value of FL and the more realistic modeling scenarios of HG, OpenFGL will inspire more users to conduct research on FHGL.

    In the future, we will continue to enhance the adaptability of OpenFGL on HGs, such as incorporating more heterogeneous graph datasets with diverse scenarios (such as PubMed~\cite{DBLP:conf/www/Xie0Y23}, Freebase~\cite{DBLP:conf/kdd/LvDLCFHZJDT21}, and OGB-MAG~\cite{hu2020ogb}), developing more advanced and efficient heterogeneous GNNs~\cite{DBLP:conf/aaai/YangYPYF23}, expanding downstream tasks (link prediction, graph classification~\cite{DBLP:conf/kdd/0015MZ0Z023}), and implementing more realistic distributed partitioning strategies~\cite{DBLP:conf/ijcai/0004K23}. 
    %We believe that in FGL, heterogeneous graphs will present more complex challenges and diversified scenarios to be addressed.%

\newpage
\newpage
\balance{
\bibliographystyle{ACM-Reference-Format}
\bibliography{reference}
}

% \begin{acks}
%  This work was supported by the [...] Research Fund of [...] (Number [...]). Additional funding was provided by [...] and [...]. We also thank [...] for contributing [...].
% \end{acks}

%\clearpage
\end{document}